\documentclass{article}

\PassOptionsToPackage{numbers}{natbib}
\usepackage[preprint]{neurips_2026}

\usepackage[utf8]{inputenc} % allow utf-8 input
\usepackage[T1]{fontenc}    % use 8-bit T1 fonts
\usepackage{hyperref}       % hyperlinks
\usepackage{url}            % simple URL typesetting
\usepackage{booktabs}       % professional-quality tables
\usepackage{amsfonts}       % blackboard math symbols
\usepackage{nicefrac}       % compact symbols for 1/2, etc.
\usepackage{microtype}      % microtypography
\usepackage{xcolor}         % colors
\usepackage{multirow}
\usepackage{graphicx}
\usepackage{subcaption}
\usepackage{amsmath}

\usepackage{algorithm}
\usepackage{algpseudocode}

\usepackage{listings}
\lstdefinestyle{mystyle}{
    commentstyle=\color{codegreen},
    keywordstyle=\color{magenta},
    numberstyle=\tiny\color{codegray},
    stringstyle=\color{codepurple},
    basicstyle=\ttfamily\footnotesize,
    breakatwhitespace=false,         
    breaklines=true,                 
    captionpos=b,                    
    keepspaces=true,                 
    numbers=left,                    
    numbersep=5pt,                  
    showspaces=false,                
    showstringspaces=false,
    showtabs=false,                  
    tabsize=2
}

\definecolor{codegreen}{rgb}{0,0.6,0}
\definecolor{codegray}{rgb}{0.5,0.5,0.5}
\definecolor{codepurple}{rgb}{0.58,0,0.82}
\definecolor{backcolour4Input}{rgb}{0.85,0.95,1}
\definecolor{backcolour4Output}{rgb}{1,0.85,0.95}

\definecolor{backcolour4Prompt}{rgb}{0.95,0.95,0.95}

\lstdefinestyle{promptstyle}{
    backgroundcolor=\color{backcolour4Prompt},
    basicstyle=\ttfamily\footnotesize,
    breaklines=true,
    breakatwhitespace=false,
    keepspaces=true,
    showspaces=false,
    showstringspaces=false,
    showtabs=false,
    numbers=none,
    frame=single,
    rulecolor=\color{gray},
    tabsize=2
}

\title{From Hazard Functions to Language Space:\\ Cox-Supervised Distillation of Survival Risk\\ into a Large Language Model}

\author{Nicholas I-Hsien Kuo$^{1}$, \textbf{Blanca Gallego}$^{1}$,
\textbf{Louisa Jorm}$^{1}$,\\ 
$^{1}$Centre for Big Data Research in Health, the University of New South Wales, Sydney, Australia\\
\footnotesize{\textcolor{white}{*}}\\
Corresponding author: Nicholas I-Hsien Kuo (\texttt{n.kuo@unsw.edu.au})
}

\begin{document}

\maketitle

\begin{abstract}
We investigate whether information about time-to-event risk estimated by a Cox proportional hazards model can be transferred into a generative large language model. We propose a text-based survival modelling pipeline in which structured clinical covariates are converted into text prompts and a Qwen-based large language model is fine-tuned to generate patient-specific survival risk using Cox model predictions as a training target. Across GBSG2, ACTG320, and WHAS500, the model achieves competitive held-out discrimination and calibration despite being trained as a text-generation task rather than with a conventional survival-analysis loss. We further analyse the geometry of the model's hidden states, where t-SNE visualisations reveal smooth risk gradients in latent space, suggesting that the model represents survival risk as a continuous structure rather than isolated risk categories. Together, these findings suggest that large language models can internalise survival-risk structure while supporting calibrated prediction, providing a route towards time-to-event reasoning in language models.
\end{abstract}

\begin{figure}[h!]
    \centering
    \includegraphics[width=\linewidth]{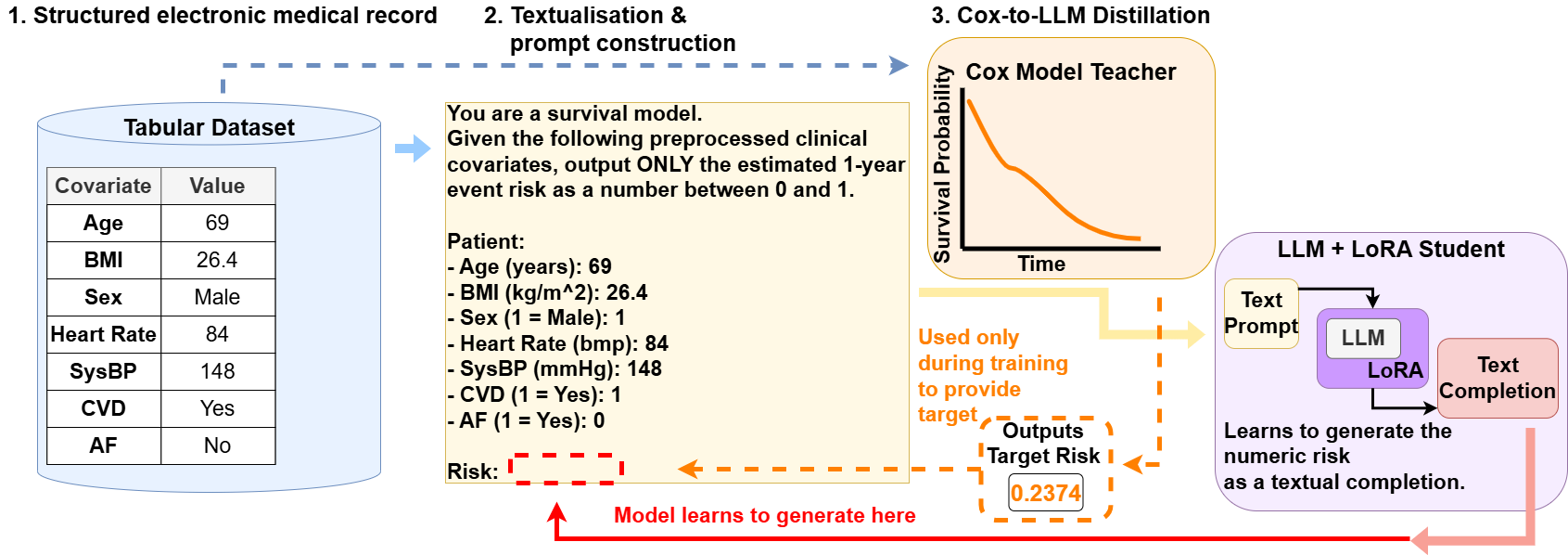}
    \caption{Text-based survival risk prediction via Cox-supervised language model training.}
    \label{fig:GraphicalAbstract}
\end{figure}

%###===###%###===###%###===###%###===###%###===###
%###===###%###===###%###===###%###===###%###===###
%###===###%###===###%###===###%###===###%###===###
%\newpage
\section{Introduction}\label{Sec:Introduction}

Survival analysis~\cite{kalbfleisch2002statistical} is a core clinical prediction problem because it models time-to-event outcomes under censoring. While classical methods such as Cox proportional hazards~\cite{cox1972regression} remain strong and interpretable, most neural survival models (\textit{e.g.,} DeepHit~\cite{lee2018deephit}) still rely on explicit predictive heads rather than language-based supervision. 

In this paper, we investigate whether time-to-event risk information estimated by a Cox model~\cite{hinton2015distilling} can be transferred into a generative large language model (LLM). As illustrated in Figure~\ref{fig:GraphicalAbstract}, structured clinical covariates are converted into natural-language prompts, while a Cox model provides patient-specific risk targets. A Qwen-based LLM~\cite{qwen2,qwen2.5} is fine-tuned through supervised text completion~\cite{brown2020language} to generate Cox-derived survival risk as text. Across GBSG2~\cite{schumacher1994randomized}, ACTG320~\cite{hammer1997controlled}, and WHAS500~\cite{goldberg1988incidence}, the model achieves competitive discrimination~\cite{harrell1982evaluating} and calibration~\cite{van2019calibration}. These findings suggest that LLMs can learn clinically meaningful time-to-event risk patterns from text-formatted clinical covariates under Cox-model supervision.

%###===###%###===###%###===###%###===###%###===###
%###===###%###===###%###===###%###===###%###===###
%###===###%###===###%###===###%###===###%###===###
%\newpage
\section{Related work}\label{Sec:RelatedWork}
Deep survival models such as DeepSurv~\cite{katzman2018deepsurv}, DeepHit~\cite{lee2018deephit}, and transformer-based clinical architectures (\textit{e.g.,} T-Risk~\cite{rao2026transformer}) learn prognostic functions directly from structured covariates or event sequences. Given patient features $x$, these models estimate a risk score or hazard function,
\(
s_\theta(x)\in\mathbb{R}
\text{ or }
h_\theta(t\mid x),
\)
and optimise survival objectives such as the Cox partial likelihood:
\begin{equation}\label{Eq:surv}
\mathcal{L}_{\mathrm{surv}}(\theta)
=
-\sum_{i:\delta_i=1}
\left[
s_\theta(x_i)
-
\log \sum_{j\in\mathcal{R}(t_i)}
e^{s_\theta(x_j)}
\right].
\end{equation}
These networks map clinical features directly to risk through a survival-predictive architecture.

Recent work on tabular large language models, including TabLLM~\cite{hegselmann2023tabllm}, reformulates structured prediction by converting rows of structured data into natural-language prompts. A structured input $x$ is transformed into text via a prompt function $\phi(x)$, and prediction is performed through
\(
p_\theta(y \mid \phi(x)).
\)
Training then uses a standard supervised objective,
\begin{equation}\label{Eq:SFT}
    \mathcal{L}_{\mathrm{cls}}(\theta)
=
-\sum_{i=1}^{n}
\log p_\theta(y_i \mid \phi(x_i)).
\end{equation} 
These approaches treat tabular prediction as language understanding, but generally focus on classification and few-shot learning rather than censored survival modelling and risk estimation. 

In this paper, we investigate whether a fine-tuned LLM trained using textual supervised fine-tuning (SFT) (Eq.~\eqref{Eq:SFT}) can approximate models trained with explicit survival objectives (Eq.~\eqref{Eq:surv}). We also examine whether clinically meaningful survival-risk patterns emerge within the model's latent space.

%###===###%###===###%###===###%###===###%###===###
%###===###%###===###%###===###%###===###%###===###
%###===###%###===###%###===###%###===###%###===###
%\newpage
\section{Methods}\label{Sec:Methods}

\begin{table}[h!]
\footnotesize
\centering
\caption{\label{Tab:WHAS500}Descriptive characteristics of the WHAS500 cohort used throughout this study.}
\begin{tabular}{|p{5.2cm}|p{5.8cm}|}
\hline
\textbf{Variable} & \textbf{Summary ([Mean ± Std Dev] or [Proportions \%])} \\
\hline
\hline
Age (years) &
$69.84 \pm 14.36$\\
\hline

Female Sex &
$38.8\%$ \\
\hline

BMI (kg/m$^2$) &
$26.45 \pm 4.15$\\
\hline

Heart Rate (bpm) &
$83.61 \pm 19.06$\\
\hline

Systolic Blood Pressure (mmHg) &
$149.16 \pm 35.85$\\
\hline

Diastolic Blood Pressure (mmHg) &
$76.28 \pm 30.94$\\
\hline

History of Cardiovascular Disease (CVD) &
$80.2\%$ \\
\hline

History of Atrial Fibrillation (AF) &
$11.8\%$ \\
\hline

History of Congestive Heart Failure (CHF) &
$29.6\%$ \\
\hline

Recurrent Myocardial Infarction &
$31.6\%$ \\
\hline

Q-wave Myocardial Infarction &
$29.4\%$ \\
\hline

Observed Event Rate &
$43.0\%$ \\
\hline

Median Follow-up Duration (days) &
$631.50$ \\
\hline
\end{tabular}
\end{table}

\paragraph{Datasets.} Experiments were conducted on three survival datasets: GBSG2~\cite{schumacher1994randomized}, ACTG320~\cite{hammer1997controlled}, and WHAS500~\cite{goldberg1988incidence}. WHAS500 consists of clinical covariates, follow-up durations, and event indicators for patients followed after acute myocardial infarction (MI). A descriptive summary of the WHAS500 cohort is provided in Table~\ref{Tab:WHAS500}; with additional details in Appendix~\S~\ref{App:MoreDetailsWHAS500}. See Appendix \S~\ref{App:MoreExperiments} for additional descriptions for GBSG2 and ACTG320. These datasets are publicly accessible through the \texttt{scikit-survival}~\cite{polsterl2020scikit} and \texttt{lifelines}~\cite{Davidson-Pilon2019} Python packages~\cite{vanRossum1995}. 

\newpage
\paragraph{Cox-supervised language model training.} Let
\(
\mathcal{D}
=
\{(x_i,t_i,\delta_i)\}_{i=1}^{N}
\)
denote a survival dataset with patient covariates \(x_i\), follow-up times \(t_i\), and event indicators \(\delta_i\). A Cox model was first fitted on the training subset to generate patient-specific survival-risk estimates \(r_i = 1 - S_i(\tau)\) at prediction horizon \(\tau\). These Cox-derived risks served as training targets for a generative language model. Rather than directly optimising a survival-specific objective, the language model learned a mapping
\(
\phi(x_i)
\rightarrow
r_i,
\)
where \(\phi(x_i)\) denotes a textual representation of patient covariates. See details in Appendix~\S~\ref{App:RiskCreation}.

\paragraph{Prompt-based survival modelling.} Each patient record was converted into a structured natural-language prompt containing the clinical covariates followed by a ``\texttt{Risk:}'' cue (Figure~\ref{fig:GraphicalAbstract}). The target completion was represented as a scalar risk string, such as:
\(
y_i = \texttt{"0.2374"}.
\)

This reformulates survival prediction as autoregressive conditional generation:
\(
p_{\theta}(y_i \mid \phi(x_i)),
\)~rather\\
than direct regression through a predictive survival head (Section \ref{Sec:RelatedWork}). Prediction is therefore mediated through tokenisation, instruction-following, and next-token generation. See details in Appendix~\S~\ref{App:CreatePrompt}.

\paragraph{LoRA fine-tuning.} We used Qwen2.5-1.5B-Instruct~\cite{qwen2.5} as the student model and fine-tuned using low-rank adaptation (LoRA)~\cite{hu2022lora}. Training was performed on prompt--completion pairs generated from the Cox teacher model:
\(
\mathcal{D}_{\mathrm{SFT}}
=
\{(\phi(x_i), y_i)\}_{i=1}^{N}.
\)
The optimisation objective was autoregressive conditional likelihood:
\(
\mathcal{L}_{\mathrm{SFT}}(\theta)
=
-
\sum_{i=1}^{N}
\log p_{\theta}(y_i \mid \phi(x_i)).
\) See details in Appendix~\S~\ref{App:LoRACompletiongSFT}.

\paragraph{Experimental evaluation.} During inference, the fine-tuned Qwen model autoregressively generated textual survival-risk completions from held-out patient prompts. The generated text was then parsed using a regular-expression-based extraction function to recover the predicted risk as a numeric scalar value. See details in Appendix~\S~\ref{App:TestTime}.

\paragraph{Experimental metrics.} Performance was evaluated using standard survival-analysis and epidemiological metrics, including Harrell's concordance index (C-index)~\cite{harrell1982evaluating}, absolute calibration error derived from the calibration slope~\cite{kuo2024ck4gen}, percentile-based net reclassification improvement (NRI)~\cite{mckearnan2018performance}, and subgroup calibration analysis. See details in Appendix~\S~\ref{App:ModelEvaluation}.

\paragraph{Code availability.}
Core implementation details are provided in the appendix, including reproducibility information (Appendix~\S~\ref{App:Reproducibility}), pipeline functions (Appendix~\S~\ref{App:PipelineOverview}), and hyperparameter settings (Appendix~\S~\ref{App:Hyperparameters}). These details are sufficient to reproduce the reported experiments. The full codebase will be released upon formal publication.

%###===###%###===###%###===###%###===###%###===###
%###===###%###===###%###===###%###===###%###===###
%###===###%###===###%###===###%###===###%###===###
%\newpage
\section{Results}\label{Sec:Results}

\begin{table}[h!]
\centering
\footnotesize
\caption{
Comparison of the Cox teacher and Qwen-based LLM student on WHAS500.}
\label{tab:whas500_combined}
\begin{tabular}{lcc}
\toprule
\multicolumn{3}{c}{\textbf{Cohort-level performance}} \\
\midrule
Metric & Cox & Qwen \\
\midrule
C-index $\uparrow$ &
$0.7658 \pm 0.0099$ &
$0.7563 \pm 0.0121$ \\

Calibration error (D21) $\downarrow$ &
$0.0583 \pm 0.0368$ &
$0.0711 \pm 0.0238$ \\

NRI$_{\text{total}}$ $\uparrow$ &
-- &
$-0.3032 \pm 0.0951$ \\

\quad NRI$_{\text{cases}}$ &
-- &
$-0.2067 \pm 0.0689$ \\

\quad NRI$_{\text{controls}}$ &
-- &
$-0.0964 \pm 0.0468$ \\

\midrule
\multicolumn{3}{c}{\textbf{Subgroup calibration error (D21) $\downarrow$}} \\
\midrule

AF &
$0.206 \pm 0.167$ &
$0.191 \pm 0.116$ \\

CHF &
$0.096 \pm 0.059$ &
$0.104 \pm 0.059$ \\

CVD &
$0.054 \pm 0.044$ &
$0.103 \pm 0.047$ \\

MI order &
$0.092 \pm 0.075$ &
$0.147 \pm 0.096$ \\

MI type &
$0.213 \pm 0.151$ &
$0.156 \pm 0.111$ \\

Sex (male) &
$0.210 \pm 0.134$ &
$0.209 \pm 0.144$ \\

\bottomrule
\end{tabular}
\end{table}

Table~\ref{tab:whas500_combined} summarises the held-out model performance on WHAS500 (reported in mean $\pm$ standard deviation). The Qwen-based LLM achieved discrimination close to that of the Cox model, with a mean C-index of $0.7563$ versus $0.7658$ for the Cox model while maintaining modestly higher calibration error ($0.0711$ versus $0.0583$). However, percentile-based NRI analysis indicates that the LLM did not fully preserve the risk ranking produced by the Cox model, with larger degradation observed for event cases (NRI${\text{cases}} = -0.2067$) than controls (NRI${\text{controls}} = -0.0964$). Subgroup calibration analysis shows a broadly similar pattern overall, although in some subgroups, such as AF and MI type, the LLM achieved slightly lower D21 calibration error than the Cox model.

\begin{figure}[t]
    \centering
    \includegraphics[width=0.4\linewidth]{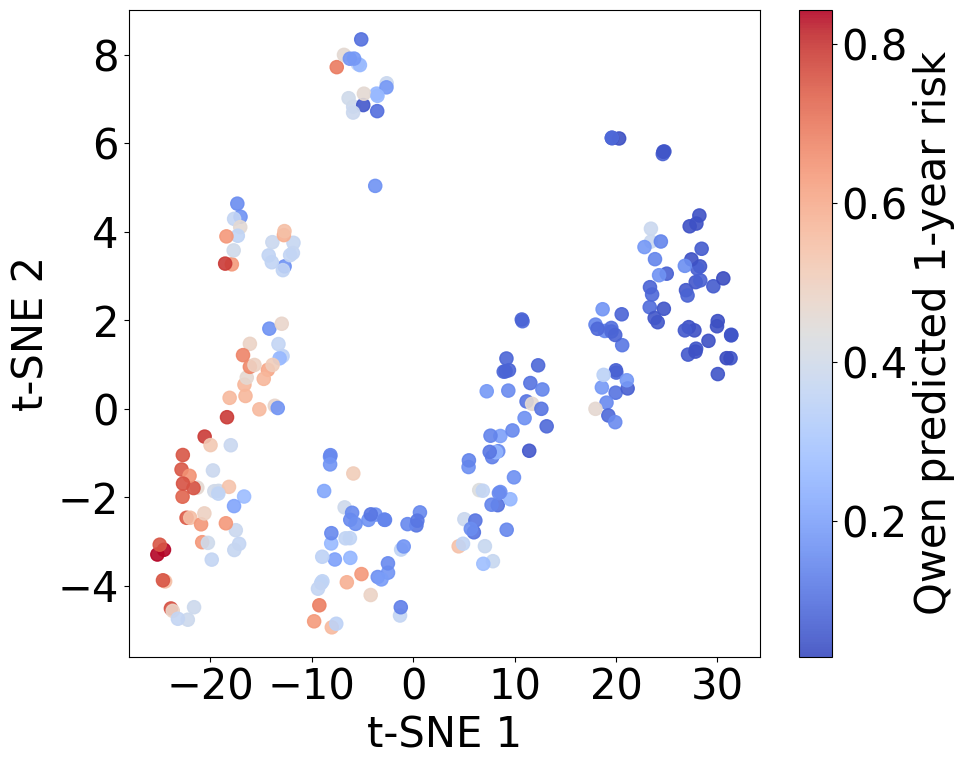}
    \caption{t-SNE projection of Qwen hidden states on the WHAS500 test set}
    \label{fig:whas500_tsne}
\end{figure}

Figure~\ref{fig:whas500_tsne} shows a two-dimensional t-SNE~\cite{van2008visualizing} projection of the final-layer hidden-state embeddings extracted from held-out WHAS500 patient prompts and coloured according to the generated 1-year survival risk. The latent representation exhibits a smooth risk gradient. This pattern suggests that the fine-tuned LLM represents predicted survival risk as a continuous structure in laten space, rather than as a set of discrete risk categories. See details of the figure setup in Appendix~\S~\ref{App:Rep}.

Additional results for GBSG2 and ACTG320 are provided in Appendix~\ref{App:MoreResults}.

%###===###%###===###%###===###%###===###%###===###
%###===###%###===###%###===###%###===###%###===###
%###===###%###===###%###===###%###===###%###===###
%\newpage
\section{Discussion}\label{Sec:Discussion}

\paragraph{Survival-risk structure in language space.}
The results suggest that survival-risk information estimated by a statistical survival model can be learned by a generative large language model through autoregressive supervised fine-tuning alone. Evaluated entirely on held-out patients, the Qwen model achieves competitive discrimination and calibration despite the absence of an explicit survival objective such as Cox partial likelihood optimisation. The smooth latent risk organisation observed in hidden-state space further suggests that the model learned internal representations that reflect clinically meaningful gradients in predicted time-to-event risk. 

\paragraph{Implications for clinical data harmonisation.}
While knowledge distillation did not help the LLM to outperform the Cox model (Table~\ref{tab:whas500_combined}), the framework may still offer practical advantages when clinical datasets are heterogeneous or difficult to standardise. Epidemiological data are frequently collected across institutions and electronic health-record systems using incompatible extraction pipelines and differing feature conventions~\cite{bouckley2025assessment}, often requiring substantial preprocessing and imputation prior to conventional survival modelling~\cite{nicholas2025estimating}. These challenges become more pronounced when relevant variables are unavailable or legally restricted across healthcare systems. For example, in cardiovascular disease research, New Zealand's PREDICT~\cite{pylypchuk2018cardiovascular} and the United Kingdom's QRISK3~\cite{hippisley2017development} explicitly incorporate ethnicity, whereas the American Heart Association's PREVENT equation~\cite{khan2024development} avoids ethnicity-specific adjustment. Such differences complicate model transferability and international standardisation. A text-based survival framework may therefore offer greater flexibility by conditioning on natural-language clinical descriptions rather than rigid feature schemas.

\paragraph{Limitations and future work.}
Several limitations remain. Since the Cox model provides the supervisory risk targets, it defines the predictive ceiling of the LLM in this study. Fine-tuning a large language model also incurs substantially greater computational cost than fitting a conventional Cox model. Moreover, the feasibility of deploying a purely text-based survival framework across heterogeneous institutions remains unverified. Future work should therefore examine robustness to informative missingness~\cite{groenwold2020informative}, heterogeneous prompt structures, and distribution shifts through systematic multi-centre ablation studies~\cite{tranchellini2026evaluating}.

%%%===%%%%%%===%%%%%%===%%%%%%===%%%%%%===%%%%%%===%%%%%%===%%%%%%===%%%%%%===%%%
\newpage
\bibliographystyle{unsrtnat}
\bibliography{A001_References}

%%%===%%%%%%===%%%%%%===%%%%%%===%%%%%%===%%%%%%===%%%%%%===%%%%%%===%%%%%%===%%%
%%%===%%%%%%===%%%%%%===%%%%%%===%%%%%%===%%%%%%===%%%%%%===%%%%%%===%%%%%%===%%%
%%%===%%%%%%===%%%%%%===%%%%%%===%%%%%%===%%%%%%===%%%%%%===%%%%%%===%%%%%%===%%%
\appendix
%%%===%%%%%%===%%%%%%===%%%%%%===%%%%%%===%%%%%%===%%%%%%===%%%%%%===%%%%%%===%%%
%%%===%%%%%%===%%%%%%===%%%%%%===%%%%%%===%%%%%%===%%%%%%===%%%%%%===%%%%%%===%%%
%%%===%%%%%%===%%%%%%===%%%%%%===%%%%%%===%%%%%%===%%%%%%===%%%%%%===%%%%%%===%%%
\newpage
\section*{Appendix: Additional Details to the Main Text}

\paragraph{Purpose of this Appendix.}
This appendix formalises the Cox-to-Qwen survival distillation framework, documents the experimental pipeline, and presents extended analyses and additional datasets used to evaluate the proposed language-model-based survival prediction approach.

\paragraph{Contents.}
\begin{itemize}

    \item \textbf{Reproducibility}
    \begin{itemize}
        \item \S~\ref{App:Reproducibility} Reproducibility Statement \dotfill \pageref{App:Reproducibility}
    \end{itemize}

    \item \textbf{Dataset Details}
    \begin{itemize}
        \item \S~\ref{App:MoreDetailsWHAS500} WHAS500 Dataset \dotfill \pageref{App:MoreDetailsWHAS500}
        \item \S~\ref{App:MoreDetailsGbsg2} GBSG2 Dataset \dotfill \pageref{App:MoreDetailsGbsg2}
        \item \S~\ref{App:MoreDetailsACTG320} ACTG320 Dataset \dotfill \pageref{App:MoreDetailsACTG320}
    \end{itemize}

    \item \textbf{Pipeline and Framework Details}
    \begin{itemize}
        \item \S~\ref{App:PipelineOverview} Experimental Pipeline \dotfill \pageref{App:PipelineOverview}
        \item \S~\ref{App:RiskCreation} Cox-derived Risk Targets \dotfill \pageref{App:RiskCreation}
        \item \S~\ref{App:CreatePrompt} Prompt Construction \dotfill \pageref{App:CreatePrompt}
        \item \hspace{5mm} \S~\ref{App:PromptTaskInstruction} Task Instruction \dotfill \pageref{App:PromptTaskInstruction}
        \item \hspace{5mm} \S~\ref{App:PromptStructuredRepresentation} Structured Patient Representation \dotfill \pageref{App:PromptStructuredRepresentation}
        \item \hspace{5mm} \S~\ref{App:PromptRiskCompletion} Risk Completion Target \dotfill \pageref{App:PromptRiskCompletion}
        \item \S~\ref{App:LoRACompletiongSFT} LoRA Fine-tuning \dotfill \pageref{App:LoRACompletiongSFT}
        \item \hspace{5mm} \S~\ref{App:LoRALoading} Base Model Loading \& LoRA Attachment \dotfill \pageref{App:LoRALoading}
        \item \hspace{5mm} \S~\ref{App:SFTOperation} Completion-only SFT \dotfill \pageref{App:SFTOperation}
        \item \hspace{5mm} \S~\ref{App:PromptRiskRange} Risk-range Bounding \dotfill \pageref{App:PromptRiskRange}
        \item \S~\ref{App:TestTime} Held-out Inference \dotfill \pageref{App:TestTime}
        \item \hspace{5mm} \S~\ref{App:HeldOutInferenceGeneration} Autoregressive Risk Generation \dotfill \pageref{App:HeldOutInferenceGeneration}
        \item \hspace{5mm} \S~\ref{App:GeneratedTextDecoding} Generated-text Decoding \dotfill \pageref{App:GeneratedTextDecoding}
        \item \S~\ref{App:ModelEvaluation} Model Evaluation \dotfill \pageref{App:ModelEvaluation}
    \end{itemize}

    \item \textbf{Representation Analysis}
    \begin{itemize}
        \item \S~\ref{App:Rep} Hidden-state Representation Analysis \dotfill \pageref{App:Rep}
    \end{itemize}

    \item \textbf{Hyperparameters}
    \begin{itemize}
        \item \S~\ref{App:Hyperparameters} Experimental Hyperparameters \dotfill \pageref{App:Hyperparameters}
    \end{itemize}

    \item \textbf{Additional Results}
    \begin{itemize}
        \item \S~\ref{App:MoreExperiments} Additional Dataset Experiments \dotfill \pageref{App:MoreExperiments}
    \end{itemize}

\end{itemize}

\paragraph{Reproducibility.}
Specifies the hardware environment, software dependencies, deterministic execution settings, and runtime configuration used throughout all experiments.

\paragraph{Dataset Details.}
Provides detailed descriptions of the WHAS500, GBSG2, and ACTG320 datasets, including preprocessing decisions, Cox modelling results, and descriptive cohort statistics.

\paragraph{Pipeline and Framework Details.}
Defines the complete Cox-to-Qwen survival distillation framework, including Cox-derived risk generation, prompt construction, LoRA-based supervised fine-tuning, held-out autoregressive inference, and survival-model evaluation procedures.

\paragraph{Representation Analysis.}
Describes the extraction and analysis of transformer hidden-state representations, including patient-level embedding construction and dimensionality-reduction analysis.

\paragraph{Hyperparameters.}
Summarises all key experimental hyperparameters and implementation settings required to reproduce the proposed framework, including data splitting, prompt formatting, LoRA configuration, supervised fine-tuning parameters, and inference settings.

\paragraph{Additional Results.}
Presents supplementary quantitative analyses on GBSG2 and ACTG320.

%###===###%###===###%###===###%###===###%###===###
%###===###%###===###%###===###%###===###%###===###
%###===###%###===###%###===###%###===###%###===###
\newpage
\section{Reproducibility Statement}\label{App:Reproducibility}

All experiments were conducted in Python 3.12.13. Random seeds were fixed across Python, NumPy, and PyTorch. CUDA deterministic algorithms were enabled where supported.

\begin{table*}[h!]
\centering
\caption{Hardware environment, runtime environment, and software dependencies.}
\label{tab:reproducibility_summary}
\small
\begin{tabular}{lll}
\toprule
\textbf{Category} & \textbf{Component} & \textbf{Specification / Version} \\
\midrule

\multirow{3}{*}{Deterministic execution}
& Deterministic algorithms & \texttt{torch.use\_deterministic\_algorithms} \\
& cuDNN deterministic & True \\
& cuDNN benchmark & False \\

\midrule

\multirow{2}{*}{System hardware}
& System RAM & 12.67 GB \\
& Available RAM at runtime & 9.18 GB \\

\midrule

\multirow{2}{*}{GPU environment}
& GPU model & NVIDIA Tesla T4 \\
& GPU memory & 15.36 GB \\

\midrule

\multirow{4}{*}{Core runtime}
& Python~\cite{vanRossum1995} & 3.12.13 \\
& PyTorch~\cite{paszke2019pytorch} & 2.10.0+cu128 \\
& cuDNN & 91002 \\

\midrule

\multirow{7}{*}{ML/DL libraries}
& transformers~\cite{wolf2019huggingface} & 5.5.0 \\
& unsloth~\cite{hanunsloth} & 2026.5.2 \\
& trl~\cite{vonwerra2020trl} & 0.24.0 \\
& peft~\cite{mangrulkar2022peft} & 0.19.1 \\
& accelerate~\cite{accelerate} & 1.13.0 \\
& bitsandbytes~\cite{dettmers2023qlora} & 0.49.2 \\
& datasets~\cite{lhoest2021datasets} & 4.3.0 \\

\midrule

\multirow{3}{*}{Statistical libraries}
& lifelines~\cite{Davidson-Pilon2019} & 0.30.3 \\
& scikit-survival~\cite{polsterl2020scikit} & 0.27.0 \\
& scikit-learn~\cite{pedregosa2011scikit} & 1.8.0 \\

\midrule

\multirow{3}{*}{Scientific libraries}
& numpy~\cite{harris2020array} & 2.0.2 \\
& pandas~\cite{mckinney2010data} & 2.2.2 \\
& matplotlib~\cite{hunter2007matplotlib} & 3.10.0 \\

\bottomrule
\end{tabular}
\end{table*}

To maximise experimental reproducibility, all experiments were executed under a deterministic PyTorch configuration with fixed random seeds across Python, NumPy, and CUDA-enabled PyTorch operations. Multiple sources of stochasticity commonly encountered in deep learning pipelines were explicitly controlled, including deterministic Python hashing via \texttt{PYTHONHASHSEED}, synchronised random seed initialisation across Python libraries, and consistent CUDA initialisation through \texttt{torch.cuda.manual\_seed()} and \texttt{torch.cuda.manual\_seed\_all()}. In addition, deterministic cuDNN kernels were enforced using \texttt{torch.backends.cudnn.deterministic=True}. While exact bitwise reproducibility cannot always be guaranteed for GPU-based deep learning workloads, these measures were implemented to maximise experimental consistency and reproducibility across repeated runs.

%%%===%%%%%%===%%%%%%===%%%%%%===%%%%%%===%%%%%%===%%%%%%===%%%%%%===%%%%%%===%%%
%%%===%%%%%%===%%%%%%===%%%%%%===%%%%%%===%%%%%%===%%%%%%===%%%%%%===%%%%%%===%%%
%%%===%%%%%%===%%%%%%===%%%%%%===%%%%%%===%%%%%%===%%%%%%===%%%%%%===%%%%%%===%%%
\newpage
\section{WHAS500 Dataset}
\label{App:MoreDetailsWHAS500}

The Worcester Heart Attack Study (WHAS) was introduced by Goldberg \textit{et al.}~\cite{goldberg1988incidence} as a longitudinal, population-based study investigating incidence and survival outcomes following acute myocardial infarction (AMI). The broader WHAS programme spans multiple observational cohorts collected between 1975 and 2001 and contains more than 11,000 hospital admissions. Due to its clinically meaningful cardiovascular covariates and time-to-event outcomes, WHAS has become a widely used benchmark dataset in survival analysis research and methodological development.

In this work, we utilise the WHAS500 subset available through the \texttt{sksurv} package~\cite{polsterl2020scikit}. WHAS500 contains 500 patients and is frequently used for demonstrations of Cox proportional hazards modelling~\cite{cox1972regression} and survival analysis methodologies~\cite{hosmer2008applied}. The dataset includes demographic information, haemodynamic measurements, cardiovascular history, myocardial infarction characteristics, follow-up duration, and survival status. These variables provide a compact but clinically realistic benchmark for evaluating both traditional survival models and modern machine learning approaches.

To validate our implementation, we first reproduced the UCLA Office of Advanced Research Computing (OARC) Cox proportional hazards example based on WHAS500. The UCLA example applies several preprocessing transformations, including polynomial BMI terms, rescaled heart rate and diastolic blood pressure variables, interaction terms between age and sex, and conversion of follow-up duration from days to years. Interested readers should search for “Table 6.7 on page 198” on the OARC website of\\
\textcolor{pink}{https://stats.oarc.ucla.edu/sas/examples/asa2/}. 

Unlike the UCLA example, our experiments employ a broader set of minimally transformed variables to increase modelling complexity and better reflect realistic clinical heterogeneity. Specifically, we include age, body mass index (BMI), sex, heart rate, systolic blood pressure, diastolic blood pressure, cardiovascular disease history, atrial fibrillation history, congestive heart failure history, myocardial infarction order, myocardial infarction type, follow-up duration, and survival status. Rather than applying extensive feature engineering or manually constructed interaction terms, we retain the original clinical variables to preserve interpretability and maintain consistency with downstream language-model-based survival prediction tasks.

Hazard ratios obtained from the expanded Cox proportional hazards model are presented in Table~\ref{Tab:AppWHAS-2}. Several clinically plausible associations were observed. Increasing age and elevated heart rate were associated with increased mortality risk, whereas higher BMI and higher diastolic blood pressure were associated with slightly lower hazards. Congestive heart failure demonstrated the strongest adverse prognostic effect, with a hazard ratio exceeding 2.0. Other variables, including prior cardiovascular disease, atrial fibrillation, myocardial infarction order, and myocardial infarction type, showed weaker associations within this cohort but were retained to preserve broader clinical context.

Descriptive statistics for the cohort are summarised in Table~\ref{Tab:WHAS500}. The cohort had a mean age of approximately 70 years and a mean BMI of 26.45 kg/m$^2$. Approximately 43\% of patients experienced the event of interest during follow-up, with a median follow-up duration of 631.5 days. Cardiovascular comorbidities were common, including congestive heart failure in 29.6\% of patients and atrial fibrillation in 11.8\% of patients. Overall, WHAS500 provides a clinically meaningful benchmark for evaluating survival analysis methods, synthetic data generation approaches, and language-model-based risk prediction systems, while also supporting transparent reproducibility and external validation against established reference implementations.

%%%===%%%%%%===%%%%%%===%%%
%%%===%%%%%%===%%%%%%===%%%
%%%===%%%%%%===%%%%%%===%%%
\newpage
\begin{table}[h!]
\footnotesize
\centering
\caption{\label{Tab:AppWHAS-2}The additional information here shows how the HR would look like in the ground truth WHAS500 dataset, if the CoxPH model were to be modelled using more variables as its inputs.}
\begin{tabular}{|p{5cm}|p{5cm}|}
\hline
\textbf{Variable} & \textbf{HR (95\% CI) (Ours)} \\
\hline
\hline
\textbf{Age} & 1.05 (1.04, 1.06) \\
\hline
\hline
\textbf{BMI} & 0.96 (0.93, 0.99) \\
\hline
\hline
\textbf{Sex} & \\
Male & Baseline \\
Female & 0.76 (0.57, 1.01) \\
\hline
\hline
\textbf{Heart Rate} & 1.01 (1.00, 1.02) \\
\hline
\textbf{Systolic Blood Pressure (SysBP)} & 1.00 (0.99, 1.01) \\
\hline
\textbf{Diastolic Blood Pressure (DiasBP)} & 0.99 (0.98, 1.00) \\
\hline
\hline
\textbf{History of\newline Cardiovascular Disease (CVD)} & \\
False & Baseline \\
True & 1.01 (0.71, 1.43) \\
\hline
\textbf{History of\newline Atrial Fibrillation (AF)} & \\
False & Baseline \\
True & 1.14 (0.81, 1.59) \\
\hline
\textbf{History of\newline Congestive Heart Failure (CHF)} & \\
False & Baseline \\
True & 2.17 (1.62, 2.91) \\
\hline
\hline
\textbf{Myocardial Infarction Order\newline (MI Order)} & \\
First & Baseline \\
Recurrent & 1.04 (0.78, 1.40) \\
\hline
\textbf{Myocardial Infarction Type\newline (MI Type)} & \\
non Q-wave & Baseline \\
Q-wave & 0.85 (0.59, 1.23) \\
\hline
\end{tabular}
\end{table}

\begin{table}[h!]
\footnotesize
\centering
\caption{\label{Tab:WHAS500}An overview of the structure and descriptive statistics for the WHAS500 synthetic dataset.}
\begin{tabular}{|p{4.25cm}|p{1.25cm}|p{3cm}|p{3.5cm}|}
\hline
\textbf{Variable} & \textbf{Type} & \textbf{Category/Statistic} & \textbf{Value/Proportion} \\
\hline
\hline
\textbf{Age} & Numeric & Mean ± Std Dev \newline Median (Q1 - Q3) & 69.84 ± 14.36 \newline 72.00 (59.00 - 82.00) \\
\hline
\hline
\textbf{BMI} & Numeric & Mean ± Std Dev \newline Median (Q1 - Q3) & 26.45 ± 4.15 \newline 25.67 (23.42 - 28.89) \\
\hline
\hline
\textbf{Sex} & Binary & Male \newline Female & Baseline \newline 38.8\% \\
\hline
\hline
\textbf{Heart Rate} & Numeric & Mean ± Std Dev \newline Median (Q1 - Q3) & 83.61 ± 19.06 \newline 80.51 (71.52 - 92.97) \\
\hline
\hline
\textbf{Systolic Blood Pressure\newline (SysBP)} & Numeric & Mean ± Std Dev \newline Median (Q1 - Q3) & 149.16 ± 35.85 \newline 146.91 (120.30 - 171.26) \\
\hline
\textbf{Diastolic Blood Pressure\newline (DiasBP)} & Numeric & Mean ± Std Dev \newline Median (Q1 - Q3) & 76.28 ± 30.94 \newline 78.86 (53.86 - 95.62) \\
\hline
\hline
\textbf{History of\newline Cardiovascular Disease (CVD)} & Binary & False \newline True & Baseline \newline 80.2\% \\
\hline
\textbf{History of\newline Atrial Fibrillation (AF)} & Binary & False \newline True & Baseline \newline 11.8\% \\
\hline
\textbf{History of\newline Congestive Heart Failure (CHF)} & Binary & False \newline True & Baseline \newline 29.6\% \\
\hline
\hline
\textbf{Myocardial Infarction Order\newline (MI Order)} & Binary & First \newline Recurrent & Baseline \newline 31.6\% \\
\hline
\textbf{Myocardial Infarction Type\newline (MI Type)} & Binary & non Q-wave \newline Q-wave & Baseline \newline 29.4\% \\
\hline
\hline
\textbf{Event} & Binary & Occurred & 43.0\% \\
\hline
\textbf{Duration} & Numeric & Mean ± Std Dev \newline Median (Q1 - Q3) & 882.44 ± 705.67 \newline 631.50 (296.50 - 1363.50) \\
\hline
\end{tabular}
\end{table}

%###===###%###===###%###===###%###===###%###===###
%###===###%###===###%###===###%###===###%###===###
%###===###%###===###%###===###%###===###%###===###
\newpage
\begin{lstlisting}[language=Python, style=mystyle, backgroundcolor=\color{backcolour4Input}]
def preprocess_whas500() -> pd.DataFrame:

    x, y = load_whas500()

    df_x = pd.DataFrame(x).rename(columns={
        "age": "Age",
        "bmi": "BMI",
        "gender": "Sex",
        "hr": "Heart Rate",
        "sysbp": "SysBP",
        "diasbp": "DiasBP",
        "cvd": "CVD",
        "afb": "AF",
        "chf": "CHF",
        "miord": "MI_Order",
        "mitype": "MI_Type",
    })

    df_y = pd.DataFrame(y).rename(columns={
        "lenfol": "Duration",
        "fstat": "Event",
    })

    df = pd.concat([df_x, df_y], axis=1)

    for col in ["Age", "BMI", "Heart Rate", "SysBP", "DiasBP", "Duration"]:
        df[col] = pd.to_numeric(df[col], errors="coerce")

    for col in ["Sex", "CVD", "AF", "CHF", "MI_Order", "MI_Type", "Event"]:
        df[col] = pd.to_numeric(df[col], errors="coerce")
        df[col] = df[col].round().astype("Int64")

    df = df.dropna(subset=[
        "Age", "BMI", "Sex", "Heart Rate", "SysBP", "DiasBP",
        "CVD", "AF", "CHF", "MI_Order", "MI_Type", "Event", "Duration"
    ]).copy()

    for col in ["Sex", "CVD", "AF", "CHF", "MI_Order", "MI_Type", "Event"]:
        df = df[df[col].isin([0, 1])].copy()
        df[col] = df[col].astype(int)

    pretty = {
        "Age": "Age (years)",
        "BMI": "BMI (kg/m^2)",
        "Sex": "Sex (1=Male)",
        "Heart Rate": "Heart Rate (bpm)",
        "SysBP": "SysBP (mmHg)",
        "DiasBP": "DiasBP (mmHg)",
        "CVD": "CVD (1=Yes)",
        "AF": "AF (1=Yes)",
        "CHF": "CHF (1=Yes)",
        "MI_Order": "MI Order (1=Recurrent)",
        "MI_Type": "MI Type (1=Q-wave)",
        "Duration": "Duration",
        "Event": "Event",
    }
    df.rename(columns=pretty, inplace=True)

    return df
\end{lstlisting}

%###===###%###===###%###===###%###===###%###===###
%###===###%###===###%###===###%###===###%###===###
%###===###%###===###%###===###%###===###%###===###
\newpage
\section{Experimental Pipeline}\label{App:PipelineOverview}

\begin{algorithm}[h!]
\caption{WHAS500 Cox-to-Qwen survival distillation pipeline}
\label{alg:whas500_pipeline}
\begin{algorithmic}[1]
\Require Dataset $D$, number of runs $R$, seed $s$, feature set $X$, event indicator $E$, duration $T$
\Ensure Run-wise metrics and saved predictions

\For{$r \leftarrow 1$ to $R$}
    \State Set random seed to $s+r-1$
    \State Split $D$ into stratified train/test sets $(D_{\text{train}}, D_{\text{test}})$ with ratio $50{:}50$ using $E$

    \State Fit a Cox proportional hazards model on $D_{\text{train}}$ using $(X, T, E)$
    \State Compute Cox 1-year risk estimates for $D_{\text{train}}$ and $D_{\text{test}}$

    \State Convert each patient row in $D_{\text{train}}$ into a prompt
    \State Pair each prompt with the corresponding Cox risk estimate as the target completion

    \State Load the 4-bit Qwen base model and attach a LoRA adapter
    \State Fine-tune the LoRA adapter on the prompt--completion pairs from $D_{\text{train}}$

    \State Switch the fine-tuned Qwen model to inference mode
    \For{each patient $i$ in $D_{\text{test}}$}
        \State Generate a risk prediction from the Qwen prompt
        \State Parse the model output as a scalar risk value
    \EndFor

    \State Evaluate Cox and Qwen on $D_{\text{test}}$ using C-index, calibration error, and percentile-NRI
    \State Save the per-run metrics, predictions, Cox model, and LoRA adapter
\EndFor

\State Aggregate metrics across runs and report mean and standard deviation
\end{algorithmic}
\end{algorithm}

The experimental pipeline consisted of multiple runs using different random seeds. For each run, the WHAS500 dataset was divided into stratified 50:50 training and testing splits based on the event indicator to preserve event prevalence across partitions. A Cox model was then fitted on the training subset using the selected clinical covariates and survival outcomes.

The fitted Cox model served as a teacher model for generating 1-year survival risk estimates. These estimated risks were computed for both the training and testing subsets and subsequently converted into prompt--completion pairs for language-model-based supervision. Each patient record was transformed into a structured natural language prompt containing the preprocessed clinical covariates, while the corresponding Cox-derived risk estimate was used as the target completion.

A Qwen2.5-1.5B-Instruct model quantised~\cite{qwen2, qwen2.5} in 4-bit precision was then fine-tuned using Low-Rank Adaptation (LoRA)~\cite{hu2022lora, dettmers2023qlora}. Fine-tuning was performed using supervised fine-tuning (SFT) with completion-only loss optimisation. The resulting LoRA adapters were saved separately for each experimental run.

Following training, the fine-tuned Qwen model was evaluated on the held-out test subset. Predicted risks were extracted from the generated text outputs and compared against both the observed survival outcomes and the CoxPH teacher model. Performance evaluation included concordance index (C-index)~\cite{harrell1982evaluating}, calibration error~\cite{kuo2024ck4gen}, percentile-based net reclassification improvement (NRI)~\cite{mckearnan2018performance}, and parsing success rate. Finally, performance metrics were aggregated across all multiple runs and summarised using mean and standard deviation statistics.

The core components of the proposed framework that we focus on are summarised below:

\begin{itemize}
    \item \textbf{Cox-derived risk targets:}\\A Cox model was used to generate 1-year survival risk estimates for supervision.
    
    \item \textbf{Prompt construction:}\\Each patient record was converted into a structured natural language prompt.
    
    \item \textbf{LoRA fine-tuning:}\\Qwen2.5 was fine-tuned using LoRA on the generated prompt--completion pairs.
    
    \item \textbf{Held-out inference:}\\The fine-tuned model was evaluated on unseen test patients using inference-only generation.
    
    \item \textbf{Model evaluation:}\\Predictions were compared using C-index, calibration error, and percentile-based NRI.
\end{itemize}

%###===###%###===###%###===###%###===###%###===###
%###===###%###===###%###===###%###===###%###===###
%###===###%###===###%###===###%###===###%###===###
\newpage
\subsection{Details 1: Cox-derived Risk Targets}\label{App:RiskCreation}

For each experimental run, a Cox model was first fitted on the training subset using the selected clinical covariates together with the survival duration and event indicator variables. Cox models are widely used in survival analysis because they provide a semi-parametric framework for modelling time-to-event outcomes while retaining interpretable covariate effects.

Let \(X_i \in \mathbb{R}^p\) denote the clinical covariates for patient \(i\). The Cox model specifies the hazard function as:
\[
h_i(t) = h_0(t)\exp(\eta_i),
\qquad
\eta_i = X_i^\top \beta,
\]
where \(h_0(t)\) denotes the baseline hazard function and \(\eta_i\) represents the linear predictor derived from the covariates and model coefficients. The Cox model therefore estimates relative hazard through the exponential transformation of the linear predictor.

From the estimated hazard, the corresponding survival probability can be reconstructed as:
\[
S_i(t) = [S_0(t)]^{\exp(\eta_i)},
\]
where \(S_0(t)\) denotes the baseline survival function. Predicted event risk at a predefined time horizon is then computed as:
\[
\hat{r}_i(t) = 1 - S_i(t).
\]

In this work, we define the prediction horizon as:
\[
t = 365.25 \text{ days},
\]
corresponding to a 1-year event risk prediction task. The fitted Cox model was subsequently used as a teacher model to generate patient-level 1-year survival risk estimates for both the training and testing subsets.

These risk estimates were generated using the following implementation:
\begin{lstlisting}[language=Python, style=mystyle, backgroundcolor=\color{backcolour4Input}]
def cox_risk_at_horizon(cph, X, horizon_days=HORIZON_DAYS):
    surv = cph.predict_survival_function(X, times=[horizon_days])
    risk = 1.0 - surv.iloc[0].values
    return pd.Series(risk, index=X.index)
\end{lstlisting}

The resulting Cox-derived risk values were stored as continuous supervision targets for both the training and testing subsets:
\begin{lstlisting}[language=Python, style=mystyle, backgroundcolor=\color{backcolour4Input}]
train_df, test_df = train_test_split(
    df, test_size=0.5, random_state=run_seed, stratify=df["Event"])

cox = CoxPHFitter()
cox.fit(train_df, duration_col="Duration", event_col="Event")

train_df["cox_risk"] = cox_risk_at_horizon(
    cox,
    train_df[feature_cols]
)

test_df["cox_risk"] = cox_risk_at_horizon(
    cox,
    test_df[feature_cols]
)
\end{lstlisting}

These Cox-derived 1-year risks formed the supervisory signal used for subsequent language-model-based survival distillation. By converting structured clinical covariates into calibrated survival risk estimates, the Cox model provided a statistically grounded teacher framework for supervising the downstream Qwen-based survival model.

%###===###%###===###%###===###%###===###%###===###
%###===###%###===###%###===###%###===###%###===###
%###===###%###===###%###===###%###===###%###===###
\newpage
\subsection{Details 2: Prompt construction}\label{App:CreatePrompt}

\begin{algorithm}[h!]
\caption{Prompt construction for Cox-derived survival supervision}
\label{alg:prompt_construction}
\begin{algorithmic}[1]
\Require Patient table $D$, feature set $X$, Cox-derived risk target $r$
\Ensure Prompt--completion pairs for model fine-tuning and evaluation

\For{each patient row $i \in D$}
    \State Initialise an empty prompt string
    \State Add a task instruction specifying 1-year risk prediction
    \State Add a patient header
    \For{each feature $x \in X$}
        \State Append the feature name and cleaned feature value to the prompt
    \EndFor
    \State Append a final \texttt{Risk:} cue
    \State Format the Cox-derived risk $r_i$ as a numeric completion string
    \State Store the pair $(\text{prompt}_i, \text{completion}_i)$
\EndFor

\State Return the full set of prompt--completion pairs
\end{algorithmic}
\end{algorithm}

After generating the Cox-derived 1-year survival risk targets, each patient record was transformed into a structured natural language representation suitable for language-model-based supervision. The objective of this stage was to convert tabular clinical covariates into a consistent textual format while preserving patient-level clinical information and maintaining compatibility with autoregressive language-model training.

Each prompt begins with a task instruction specifying the prediction objective, followed by a structured patient section containing the preprocessed clinical covariates and their corresponding values. Continuous variables were rendered as readable numeric values, while binary variables were retained in their cleaned integer form. The prompt concludes with a \texttt{Risk:} cue, which instructs the model to generate a scalar survival risk estimate as the completion target.

The resulting prompt--completion pairs were used as the supervision format for downstream LoRA-based fine-tuning of the Qwen survival model. Importantly, the same prompt structure was consistently applied across both training and testing subsets to minimise formatting-induced variability and preserve reproducibility across experimental runs.

The core components of the prompt construction framework that we focus on are summarised below:

\begin{itemize}
    \item \textbf{Task instruction:}\\Each prompt begins with an instruction defining the task as 1-year survival risk prediction.
    
    \item \textbf{Structured patient representation:}\\Clinical covariates were serialised into a fixed template containing feature names and cleaned feature values.
    
    \item \textbf{Risk completion target:}\\The Cox-derived 1-year survival risk was formatted as a scalar numeric completion.
    
\end{itemize}

%###===###%###===###%###===###%###===###%###===###
%###===###%###===###%###===###%###===###%###===###
%###===###%###===###%###===###%###===###%###===###
\newpage
\subsubsection{Details 2.1: Task instruction}
\label{App:PromptTaskInstruction}

The prompt begins with a fixed instruction that defines the prediction task as 1-year survival risk estimation. In the implementation, this behaviour is encoded directly in the \texttt{make\_prompt()} function, which prepends a task instruction before listing the clinical covariates. The second sentence is the key task-defining instruction, as it specifies both the output quantity and the prediction horizon. The subsequent \texttt{Patient:} marker introduces the structured covariate block and is therefore part of the input formatting rather than the instruction itself.

The relevant implementation is shown below:

\begin{lstlisting}[language=Python, style=mystyle, backgroundcolor=\color{backcolour4Input}]
def make_prompt(row):
    lines = [
        "You are a survival model.",
        "Given the following preprocessed clinical covariates, output ONLY the estimated 1-year event risk as a number between 0 and 1.",
        "",
        "Patient:",
    ]
    for col in feature_cols:
        val = row[col]
        if isinstance(val, (np.integer, int)):
            display_val = int(val)
        else:
            try:
                display_val = float(val)
                if display_val.is_integer():
                    display_val = int(display_val)
            except Exception:
                display_val = val
        lines.append(f"- {col}: {display_val}")
    lines.append("")
    lines.append("Risk:")
    return "\n".join(lines)
\end{lstlisting}

This instruction-based template is used consistently across all runs and for both the training and testing subsets. The resulting prompt clearly frames the model as a survival risk predictor and provides a fixed textual cue for generating a scalar risk completion.

An illustrative example prompt constructed from WHAS500 variables is shown below. The exact values vary by patient, but the structure remains fixed.

\begin{lstlisting}[style=promptstyle]
You are a survival model.
Given the following preprocessed clinical covariates, output ONLY the estimated 1-year event risk as a number between 0 and 1.

Patient:
- Age (years): 69
- BMI (kg/m^2): 26.4
- Sex (1=Male): 1
- Heart Rate (bpm): 84
- SysBP (mmHg): 148
- DiasBP (mmHg): 76
- CVD (1=Yes): 1
- AF (1=Yes): 0
- CHF (1=Yes): 1
- MI Order (1=Recurrent): 0
- MI Type (1=Q-wave): 1

Risk:
\end{lstlisting}

This prompt format ensures that the language model receives an explicit, standardised task instruction before the patient covariates, supporting reproducible prompt-based survival distillation.

%###===###%###===###%###===###%###===###%###===###
%###===###%###===###%###===###%###===###%###===###
%###===###%###===###%###===###%###===###%###===###
\newpage
\subsubsection{Details 2.2: Structured patient representation}
\label{App:PromptStructuredRepresentation}

The structured patient representation is implemented by first standardising the WHAS500 variables into a compact modelling frame and then serialising each patient’s covariates into a fixed text template. This ensures that the same clinical fields appear in the same order for every patient, with cleaned and human-readable feature names.

The dataset preparation step that produces prompt-friendly variables is shown below:
\begin{lstlisting}[language=Python, style=mystyle, backgroundcolor=\color{backcolour4Input}]
def preprocess_whas500():

    ...

    return df

df = preprocess_whas500()
feature_cols = [c for c in df.columns if c not in ["Duration", "Event"]]
\end{lstlisting}

Refer to more details of the \texttt{preprocess\_whas500()} function in Appendix~\S~\ref{App:MoreDetailsWHAS500}.

After preprocessing, the patient covariates are serialised into a fixed prompt structure through the \texttt{make\_prompt()} function (Appendix \S~\ref{App:PromptTaskInstruction}). The core mechanism responsible for transforming the tabular WHAS500 row into a textual representation is the iteration over \texttt{feature\_cols}, where each covariate is appended as a structured \texttt{feature: value} pair.

The relevant implementation is shown below:
\begin{lstlisting}[language=Python, style=mystyle, backgroundcolor=\color{backcolour4Input}]
for col in feature_cols:
    val = row[col]

    if isinstance(val, (np.integer, int)):
        display_val = int(val)
    else:
        try:
            display_val = float(val)
            if display_val.is_integer():
                display_val = int(display_val)
        except Exception:
            display_val = val

    lines.append(f"- {col}: {display_val}")
\end{lstlisting}

This loop ensures that every patient is represented using the same ordered feature set and formatting scheme. Continuous variables are converted into readable numeric values, while binary variables are preserved in integer form. The resulting representation maintains the original clinical information while imposing a stable and reproducible textual structure suitable for autoregressive language-model processing.

This is the loop that creates:
\begin{lstlisting}[style=promptstyle]
Patient:
- Age (years): 69
- BMI (kg/m^2): 26.4
- Sex (1=Male): 1
- Heart Rate (bpm): 84
- SysBP (mmHg): 148
- DiasBP (mmHg): 76
- CVD (1=Yes): 1
- AF (1=Yes): 0
- CHF (1=Yes): 1
- MI Order (1=Recurrent): 0
- MI Type (1=Q-wave): 1
\end{lstlisting}

of the input prompt.

%###===###%###===###%###===###%###===###%###===###
%###===###%###===###%###===###%###===###%###===###
%###===###%###===###%###===###%###===###%###===###
\newpage
\subsubsection{Details 2.3: Risk completion target}
\label{App:PromptRiskCompletion}

After generating the Cox-derived 1-year survival risks (Appendix \S~\ref{App:RiskCreation}), the resulting values were converted into scalar numeric completion targets for supervised language-model training. The objective of this step was to provide the Qwen model with a compact and deterministic target representation that could be directly optimised using completion-only supervised fine-tuning.

The formatting of the completion target is implemented in the \texttt{make\_completion()} function:
\begin{lstlisting}[language=Python, style=mystyle, backgroundcolor=\color{backcolour4Input}]
def make_completion(risk_value):
    return f"{float(risk_value):.4f}"
\end{lstlisting}

This function converts the Cox-derived risk into a fixed decimal representation with four decimal places in string format. The resulting completion therefore corresponds to a single scalar numeric value between 0 and 1, representing the estimated 1-year event risk for the associated patient.

The completion targets are then attached to the training and testing subsets as follows:
\begin{lstlisting}[language=Python, style=mystyle, backgroundcolor=\color{backcolour4Input}]
train_df["completion"] = (
    train_df["cox_risk"].apply(make_completion)
)

test_df["completion"] = (
    test_df["cox_risk"].apply(make_completion)
)
\end{lstlisting}

A corresponding prompt--completion training example therefore takes the following form:
\begin{lstlisting}[style=promptstyle]
Prompt:
You are a survival model.
Given the following preprocessed clinical covariates, output ONLY the estimated 1-year event risk as a number between 0 and 1.

Patient:
- Age (years): 69
- BMI (kg/m^2): 26.4
- Sex (1=Male): 1
- Heart Rate (bpm): 84
- SysBP (mmHg): 148
- DiasBP (mmHg): 76
- CVD (1=Yes): 1
- AF (1=Yes): 0
- CHF (1=Yes): 1
- MI Order (1=Recurrent): 0
- MI Type (1=Q-wave): 1

Risk:

Completion:
0.2374
\end{lstlisting}

By representing the target as a scalar numeric value encoded in fixed-length string format (\textit{i.e.,} the \texttt{"0.2374"} shown above), the framework preserves compatibility with autoregressive language-model training while simplifying downstream risk extraction and evaluation during inference. Importantly, the completion target is not stored as a floating-point tensor label in the conventional regression sense, but rather as a textual sequence prediction target. This design is necessary because the Qwen model operates as an autoregressive language model trained through next-token prediction, where gradients are propagated through discrete token sequences rather than continuous scalar outputs. Consequently, the Cox-derived 1-year survival risk is serialised into a four-decimal-place string representation between 0 and 1, enabling the model to learn survival-risk generation using standard language-model supervision and autograd-compatible token prediction.

This also forces the model to learn a direct mapping from purely text-based clinical covariates to purely text-based predicted survival risk.

%###===###%###===###%###===###%###===###%###===###
%###===###%###===###%###===###%###===###%###===###
%###===###%###===###%###===###%###===###%###===###
\newpage
\subsection{Details 3: LoRA fine-tuning}\label{App:LoRACompletiongSFT}

\begin{algorithm}[h!]
\caption{LoRA-based fine-tuning of Qwen on Cox-derived supervision}
\label{alg:lora_finetuning}
\begin{algorithmic}[1]
\Require Training set $D_{\text{train}}$ containing prompts and Cox-derived completions, model $M$, seed $s$
\Ensure Fine-tuned LoRA adapter and saved tokenizer

\State Load the 4-bit Qwen base model and tokenizer
\State Attach LoRA adapters to the selected transformer layers
\State Switch the model into training mode
\State Convert the prompt--completion pairs into a dataset object
\State Configure supervised fine-tuning with completion-only loss
\State Train the model on the prompt--completion pairs from $D_{\text{train}}$
\State Save the learned LoRA adapter and tokenizer
\State Return the fine-tuned model and tokenizer
\end{algorithmic}
\end{algorithm}

The next stage of the pipeline fine-tunes the Qwen2.5 base model using Low-Rank Adaptation (LoRA) on the generated prompt--completion pairs. The objective is to adapt a compact language model to the survival prediction task without updating all of the base model parameters. Instead, trainable low-rank adapter weights are inserted into selected attention and feed-forward projection layers, allowing efficient supervised adaptation under limited computational resources.

In implementation, the 4-bit quantised Qwen base model is first loaded, after which LoRA adapters are attached to the relevant projection modules. The model is then switched into training mode and optimised using supervised fine-tuning on the Cox-derived prompt--completion dataset (Appendice \S~\ref{App:PromptStructuredRepresentation}--\ref{App:PromptRiskCompletion}). Completion-only loss is used so that the model is trained specifically to generate the scalar survival-risk completion from the prompt context. After training, the resulting LoRA adapter and tokenizer are saved for subsequent held-out evaluation and inference.

The core components of the LoRA fine-tuning framework that we focus on are summarised below:

\begin{itemize}
    \item \textbf{Base model loading \& LoRA adapter attachment:}\\The Qwen2.5 model was loaded in 4-bit quantised form with the majority of weights frozen, while Low-Rank Adaptation (LoRA) modules were inserted into selected transformer projection layers.
    
    \item \textbf{Supervised fine-tuning with completion-only loss:}\\The model was trained on Cox-derived prompt--completion pairs using autoregressive supervised fine-tuning, where optimisation was performed only on the completion tokens corresponding to the predicted survival risk.
    
    \item \textbf{Risk-range instruction and post-hoc bounding:}\\The model was instructed to generate survival risks between 0 and 1 through prompt engineering and bounded supervision targets, while output clipping to \([0,1]\) was applied only during downstream evaluation.
\end{itemize}

%###===###%###===###%###===###%###===###%###===###
%###===###%###===###%###===###%###===###%###===###
%###===###%###===###%###===###%###===###%###===###
\newpage
\subsubsection{Details 3.1: Base model loading \& LoRA adapter attachment}
\label{App:LoRALoading}

To enable efficient large language model fine-tuning under limited computational resources, we utilised the Unsloth framework~\cite{hanunsloth}, which provides accelerated and memory-efficient training utilities for transformer-based models. Unsloth integrates quantised model loading, padding-free optimisation, gradient checkpointing, and parameter-efficient fine-tuning techniques such as LoRA. In this work, Unsloth was used to load the Qwen2.5-1.5B-Instruct base model in 4-bit precision and to attach trainable LoRA adapters for downstream survival-model adaptation.

The LoRA fine-tuning pipeline begins by loading the Qwen base model and then attaching trainable low-rank adaptation modules to selected transformer projection layers. This design enables parameter-efficient adaptation while keeping the majority of the original Qwen weights frozen during optimisation. Mathematically, LoRA replaces a full weight update
\[
W \leftarrow W + \Delta W
\]
with a low-rank decomposition:
\[
\Delta W = BA,
\]
where
\[
B \in \mathbb{R}^{d \times r},
\qquad
A \in \mathbb{R}^{r \times k},
\qquad
r \ll \min(d,k).
\]
Here, \(W \in \mathbb{R}^{d \times k}\) denotes the original frozen transformer weight matrix, while \(A\) and \(B\) are trainable low-rank matrices. Instead of updating the full parameter matrix \(W\), LoRA learns only the low-rank approximation \(\Delta W\), substantially reducing the number of trainable parameters and GPU memory requirements during fine-tuning.

The relevant implementation is shown below:
\begin{lstlisting}[language=Python, style=mystyle, backgroundcolor=\color{backcolour4Input}]
def load_llm():
    model, tokenizer = FastLanguageModel.from_pretrained(
        model_name=MODEL_NAME,
        max_seq_length=MAX_SEQ_LENGTH,
        load_in_4bit=True,
        dtype=None,
    )

    tokenizer.padding_side = "right"
    if tokenizer.pad_token is None:
        tokenizer.pad_token = tokenizer.eos_token

    model = FastLanguageModel.get_peft_model(
        model,
        r=16,
        target_modules=[
            "q_proj", "k_proj", "v_proj", "o_proj",
            "gate_proj", "up_proj", "down_proj",
        ],
        lora_alpha=16,
        lora_dropout=0.05,
        bias="none",
        use_gradient_checkpointing="unsloth",
        random_state=SEED,
        use_rslora=False,
    )

    return model, tokenizer
\end{lstlisting}

The base Qwen model is loaded using:
\begin{lstlisting}[language=Python, style=mystyle, backgroundcolor=\color{backcolour4Input}]
model, tokenizer = FastLanguageModel.from_pretrained(
    model_name=MODEL_NAME,
    max_seq_length=MAX_SEQ_LENGTH,
    load_in_4bit=True,
    dtype=None,
)
\end{lstlisting}

During loading, the tokeniser configuration is automatically aligned with the model configuration, including the handling of PAD/BOS/EOS tokens. The training logs indicated that the tokeniser configuration differed slightly from the original model configuration, prompting Unsloth to synchronise the corresponding settings automatically.

The LoRA adapters are subsequently attached using:
\begin{lstlisting}[language=Python, style=mystyle, backgroundcolor=\color{backcolour4Input}]
model = FastLanguageModel.get_peft_model(
    model,
    r=16,
    target_modules=[
        "q_proj", "k_proj", "v_proj", "o_proj",
        "gate_proj", "up_proj", "down_proj",
    ],
    lora_alpha=16,
    lora_dropout=0.05,
    bias="none",
    use_gradient_checkpointing="unsloth",
    random_state=SEED,
    use_rslora=False,
)
\end{lstlisting}

The selected target modules correspond to the attention and feed-forward projection layers within the transformer architecture. Training logs produced by Unsloth confirmed that only the LoRA parameters remained trainable:
\begin{lstlisting}[style=promptstyle]
Trainable parameters = 18,464,768
of 1,562,179,072 (1.18% trained)
\end{lstlisting}

This demonstrates that the majority of the Qwen base model remained frozen throughout fine-tuning, while only approximately 1.18\% of parameters were updated. The logs also indicated that Unsloth automatically enabled padding-free optimisation for faster training and successfully completed the full supervised fine-tuning process over designated epochs.

%###===###%###===###%###===###%###===###%###===###
%###===###%###===###%###===###%###===###%###===###
%###===###%###===###%###===###%###===###%###===###
\newpage
\subsubsection{Details 3.2: Supervised fine-tuning with completion-only loss}
\label{App:SFTOperation}

The LoRA-adapted Qwen model was trained using supervised fine-tuning (SFT) on the Cox-derived prompt--completion pairs. In this setting, SFT is formulated as an autoregressive conditional language-modelling objective: given a prompt \(x\) and a target completion \(y = (y_1,\dots,y_T)\), the model is optimised to maximise the conditional log-likelihood of the completion,
\[
\max_{\theta}\ \sum_{(x,y)\in\mathcal{D}} \log p_{\theta}(y \mid x),
\]
which, for an autoregressive language model, factorises token by token as
\[
\log p_{\theta}(y \mid x)
=
\sum_{t=1}^{T} \log p_{\theta}(y_t \mid x, y_{<t}).
\]
Equivalently, training minimises the negative log-likelihood,
\[
\mathcal{L}_{\mathrm{SFT}}(\theta)
=
- \sum_{(x,y)\in\mathcal{D}} \sum_{t=1}^{T} \log p_{\theta}(y_t \mid x, y_{<t}).
\]

In our implementation, the prompt is treated as conditioning context and the loss is computed only on the completion tokens. This is realised through the completion-only loss setting in the trainer configuration, which prevents the model from being penalised for reproducing the prompt itself and instead focuses optimisation entirely on generating the Cox-derived survival-risk target.

The relevant implementation is shown below:
\begin{lstlisting}[language=Python, style=mystyle, backgroundcolor=\color{backcolour4Input}]
def train_lora_adapter(train_df, output_dir, seed=SEED):
    train_df = train_df.copy()

    train_df["prompt"] = train_df.apply(make_prompt, axis=1)
    train_df["completion"] = train_df["cox_risk"].apply(make_completion)

    train_ds = Dataset.from_pandas(train_df[["prompt", "completion"]], preserve_index=False)

    model, tokenizer = load_llm()
    FastLanguageModel.for_training(model)

    sft_config = SFTConfig(
        output_dir=output_dir,
        per_device_train_batch_size=2,
        gradient_accumulation_steps=8,
        num_train_epochs=10,
        learning_rate=1e-4,
        logging_steps=10,
        save_steps=50,
        eval_strategy="no",
        report_to="none",
        fp16=True,
        bf16=False,
        seed=seed,
        max_length=MAX_SEQ_LENGTH,
        completion_only_loss=True,
        eos_token="<|im_end|>",
        remove_unused_columns=False,
    )

    trainer = SFTTrainer(
        model=model,
        args=sft_config,
        train_dataset=train_ds,
        processing_class=tokenizer,
    )

    trainer.train()
\end{lstlisting}

\underline{Lines 34, in \texttt{trainer = SFTTrainer([...])}}:\\
The key steps are the construction of the prompt--completion dataset (Appendice \S~\ref{App:PromptStructuredRepresentation}--\ref{App:PromptRiskCompletion}),
\begin{lstlisting}[language=Python, style=mystyle, backgroundcolor=\color{backcolour4Input}]
train_df["prompt"] = train_df.apply(make_prompt, axis=1)
train_df["completion"] = train_df["cox_risk"].apply(make_completion)
train_ds = Dataset.from_pandas(train_df[["prompt", "completion"]], preserve_index=False)
\end{lstlisting}
\underline{Lines 26, in \texttt{sft\_config = SFTConfig([...])} \&}\\
\underline{Lines 33, in \texttt{trainer = SFTTrainer([...])}}:\\
and the completion-only optimisation setting,
\begin{lstlisting}[language=Python, style=mystyle, backgroundcolor=\color{backcolour4Input}]
completion_only_loss=True
\end{lstlisting}
which ensures that the model is trained specifically to map the prompt context to the Cox-derived scalar risk string.

This SFT stage therefore provides the optimisation mechanism that aligns the LoRA-adapted Qwen model with the Cox teacher’s risk targets under an autoregressive next-token prediction objective.

%###===###%###===###%###===###%###===###%###===###
%###===###%###===###%###===###%###===###%###===###
%###===###%###===###%###===###%###===###%###===###
\newpage
\subsubsection{Details 3.3: Risk-range instruction and post-hoc bounding}
\label{App:PromptRiskRange}

Although the prompt-completion objective encourages the Qwen model to produce a scalar risk estimate in the form of a four-decimal string within \([0,1]\), the implementation does not impose a hard decoding constraint at generation time. Instead, the model is instructed through prompt engineering to output the estimated 1-year event risk as a number between 0 and 1 (Appendix \S~\ref{App:PromptTaskInstruction}), while the Cox-derived supervision targets are themselves formatted as bounded four-decimal strings (Appendix \S~\ref{App:PromptRiskCompletion}). In practice, across repeated testing runs, we consistently obtained valid four-decimal outputs in the unit interval, with no observed parsing failures.

The relevant generation step simply samples the model output autoregressively:
\begin{lstlisting}[language=Python, style=mystyle, backgroundcolor=\color{backcolour4Input}]
def generate_risk(model, tokenizer, prompt):
    inputs = tokenizer(prompt, return_tensors="pt").to("cuda")
    with torch.no_grad():
        outputs = model.generate(
            **inputs,
            max_new_tokens=16,
            do_sample=False,
            temperature=0.0,
        )
    prompt_len = inputs["input_ids"].shape[1]
    gen_tokens = outputs[0][prompt_len:]
    gen_text = tokenizer.decode(gen_tokens, skip_special_tokens=True).strip()
    pred = extract_first_float(gen_text)
    return gen_text, pred
\end{lstlisting}

\underline{Lines 13}:\\
The model output is then parsed from the generated text:
\begin{lstlisting}[language=Python, style=mystyle, backgroundcolor=\color{backcolour4Input}]
pred = extract_first_float(gen_text)
\end{lstlisting}

This means that the range constraint is learned and encouraged through the prompt and supervision format, rather than enforced as a decoding rule. Post-hoc clipping is only applied during evaluation:
\begin{lstlisting}[language=Python, style=mystyle, backgroundcolor=\color{backcolour4Input}]
df_eval[pred_col] = df_eval[pred_col].clip(0, 1)
nri_df[pred_col] = nri_df[pred_col].clip(0, 1)
nri_df["cox_risk"] = nri_df["cox_risk"].clip(0, 1)
\end{lstlisting}

In this way, the framework remains fully autoregressive while still producing numerically bounded survival-risk outputs in a reproducible text-to-risk format.

%###===###%###===###%###===###%###===###%###===###
%###===###%###===###%###===###%###===###%###===###
%###===###%###===###%###===###%###===###%###===###
\newpage
\subsection{Details 4: Held-out inference}\label{App:TestTime}

\begin{algorithm}[h!]
\caption{Held-out inference and risk generation on unseen test patients}
\label{alg:heldout_inference}
\begin{algorithmic}[1]
\Require Held-out test set $D_{\text{test}}$, fine-tuned LoRA adapter, tokeniser
\Ensure Generated survival risks and parsed scalar predictions

\State Load the saved Qwen base model and LoRA adapter
\State Switch the model into inference-only mode
\For{each patient row $i \in D_{\text{test}}$}
    \State Construct the structured patient prompt
    \State Tokenise the prompt and move tensors to GPU
    \State Generate autoregressive completion tokens
    \State Decode the generated text output
    \State Extract the first numeric value from the generated text
    \State Store the generated text and parsed scalar risk prediction
\EndFor

\State Return the generated predictions and parsed risk values
\end{algorithmic}
\end{algorithm}

After LoRA fine-tuning (Appendix \S~\ref{App:SFTOperation}), the resulting Qwen survival model was evaluated on unseen held-out patients using inference-only autoregressive generation (Appendix \S~\ref{App:RiskCreation}). The objective of this stage was to assess whether the fine-tuned language model could reconstruct clinically meaningful survival-risk estimates (Appendix \S~\ref{App:PromptRiskRange}) purely from structured textual patient prompts without further parameter updates.

In implementation, the saved LoRA adapter was reattached to the Qwen base model and the model was explicitly switched into inference mode before evaluation. Each patient from the held-out testing subset was converted into the same structured prompt format used during training (Appendix~\S~\ref{App:PromptTaskInstruction}--\ref{App:PromptRiskCompletion}). The model then generated a short textual completion corresponding to the predicted 1-year survival risk.

Because the model operates autoregressively, the generated output is initially produced as free-form text rather than a directly accessible scalar value (Appendix \S~\ref{App:PromptRiskRange}). Consequently, the generated completion was decoded and post-processed using lightweight parsing to recover the first valid numeric prediction from the output sequence. The resulting scalar risk estimates were subsequently used for downstream evaluation.

The core components of the held-out inference framework that we focus on are summarised below:
\begin{itemize}
    \item \textbf{Autoregressive survival-risk generation:}\\Structured patient prompts from the held-out test set were provided to the model, which generated textual survival-risk completions autoregressively.

    \item \textbf{Generated-text decoding and numeric extraction:}\\The generated text outputs were decoded and post-processed using lightweight parsing utilities to recover scalar survival-risk predictions.

\end{itemize}

%###===###%###===###%###===###%###===###%###===###
%###===###%###===###%###===###%###===###%###===###
%###===###%###===###%###===###%###===###%###===###
\newpage
\subsubsection{Details 4.1: Autoregressive survival-risk generation}
\label{App:HeldOutInferenceGeneration}

After fine-tuning, the model is evaluated on the held-out WHAS500 test split using inference-only generation: the saved LoRA adapter is reloaded, the model is switched to inference mode, and each test-time patient prompt is passed through the autoregressive decoder to generate a textual survival-risk completion.

Autoregressive risk prediction is carried out by \texttt{generate\_risk()}, which tokenises the prompt, generates new tokens without sampling, and extracts the first floating-point value from the generated text:
\begin{lstlisting}[language=Python, style=mystyle, backgroundcolor=\color{backcolour4Input}]
def generate_risk(model, tokenizer, prompt):
    inputs = tokenizer(prompt, return_tensors="pt").to("cuda")
    with torch.no_grad():
        outputs = model.generate(
            **inputs,
            max_new_tokens=16,
            do_sample=False,
            temperature=0.0,
        )
    prompt_len = inputs["input_ids"].shape[1]
    gen_tokens = outputs[0][prompt_len:]
    gen_text = tokenizer.decode(gen_tokens, skip_special_tokens=True).strip()
    pred = extract_first_float(gen_text)
    return gen_text, pred
\end{lstlisting}

The decoding is applied to each patient in the held-out test set using the same prompt format as in training:
\begin{lstlisting}[language=Python, style=mystyle, backgroundcolor=\color{backcolour4Input}]
eval_rows = []
for idx, row in test_df.iterrows():
    gen_text, pred = generate_risk(model, tokenizer, row["prompt"])
    eval_rows.append({
        "row_id": idx,
        "prompt": row["prompt"],
        "cox_risk": float(row["cox_risk"]),
        "qwen_raw_output": gen_text,
        "qwen_pred_risk": pred,
        "parse_success": pred is not None,
        "abs_error": None if pred is None else abs(pred - float(row["cox_risk"])),
    })
\end{lstlisting}

This held-out inference procedure is purely generative: the model receives a structured clinical prompt and produces a text completion token by token, from which the survival-risk estimate is parsed after decoding. In this way, the downstream prediction task remains aligned with the autoregressive language-modelling formulation used during LoRA fine-tuning, while evaluation is performed exclusively on unseen test patients.

%###===###%###===###%###===###%###===###%###===###
%###===###%###===###%###===###%###===###%###===###
%###===###%###===###%###===###%###===###%###===###
\newpage
\subsubsection{Details 4.2: Generated-text decoding and numeric extraction}
\label{App:GeneratedTextDecoding}

After inference, the model does not emit a native scalar tensor (Appendix \S~\ref{App:PromptRiskRange}). Instead, it produces a text continuation, which must be decoded and parsed to recover a numeric survival-risk prediction. This is handled by two lightweight helper functions:\\
\texttt{generate\_risk()} and \texttt{extract\_first\_float()}.

The first function performs autoregressive generation and returns both the raw decoded text and the parsed risk value:
\begin{lstlisting}[language=Python, style=mystyle, backgroundcolor=\color{backcolour4Input}]
def generate_risk(model, tokenizer, prompt):
    inputs = tokenizer(prompt, return_tensors="pt").to("cuda")
    with torch.no_grad():
        outputs = model.generate(
            **inputs,
            max_new_tokens=16,
            do_sample=False,
            temperature=0.0,
        )
    prompt_len = inputs["input_ids"].shape[1]
    gen_tokens = outputs[0][prompt_len:]
    gen_text = tokenizer.decode(gen_tokens, skip_special_tokens=True).strip()
    pred = extract_first_float(gen_text)
    return gen_text, pred
\end{lstlisting}

The second function extracts the first numeric token from the generated string:
\begin{lstlisting}[language=Python, style=mystyle, backgroundcolor=\color{backcolour4Input}]
def extract_first_float(text):
    match = re.search(r"[-+]?\d*\.?\d+", text)
    if match:
        try:
            return float(match.group())
        except ValueError:
            return None
    return None
\end{lstlisting}

Together, these functions convert the model’s free-form textual continuation into a scalar prediction that can be evaluated quantitatively. This design is useful because the model is trained as an autoregressive language model, so its outputs are naturally strings rather than typed floating-point values.

A few examples illustrate the mapping:

\begin{lstlisting}[style=promptstyle]
Generated text:
0.2374

Parsed output:
0.2374
\end{lstlisting}

\begin{lstlisting}[style=promptstyle]
Generated text:
The estimated risk is 0.2374.

Parsed output:
0.2374
\end{lstlisting}

\begin{lstlisting}[style=promptstyle]
Generated text:
Risk: 0.2374 (approximately)

Parsed output:
0.2374
\end{lstlisting}

In practice, this parsing step makes evaluation robust to minor formatting variation in the generated continuation. It also keeps the downstream pipeline simple: the raw text is preserved for inspection, while the extracted float is used for C-index, calibration, and NRI computation.

%###===###%###===###%###===###%###===###%###===###
%###===###%###===###%###===###%###===###%###===###
%###===###%###===###%###===###%###===###%###===###
\newpage
\subsection{Details 5: Model evaluation}\label{App:ModelEvaluation}

\begin{algorithm}[h!]
\caption{Evaluation of Cox and Qwen survival-risk predictions}
\label{alg:model_evaluation}
\begin{algorithmic}[1]
\Require Predicted survival risks, observed durations $T$, event indicators $E$
\Ensure C-index, calibration statistics, percentile-based NRI, and aggregated run-wise metrics

\For{each experimental run}
    \State Merge predicted risks with observed survival outcomes
    \State Remove invalid or unparsable predictions
    \State Clip predicted risks to the interval $[0,1]$

    \State Compute C-index for discrimination analysis
    \State Group predictions into quantile-based risk bins
    \State Estimate observed survival risk using Kaplan--Meier estimation
    \State Fit calibration regression and compute calibration error

    \State Compute percentile-based NRI relative to Cox teacher predictions

    \State Store run-wise evaluation metrics
\EndFor

\State Aggregate metrics across runs using mean and standard deviation
\end{algorithmic}
\end{algorithm}

After held-out inference (Appendix~\S~\ref{App:HeldOutInferenceGeneration}--\ref{App:GeneratedTextDecoding}), the generated Qwen survival-risk predictions were quantitatively evaluated against both the observed survival outcomes and the Cox teacher model predictions. The objective of this stage was to assess discrimination performance, calibration quality, and relative patient risk reclassification under multiple experimental runs.

Three primary evaluation metrics were used throughout the framework:
\begin{enumerate}
    \item Harrell's C-index for survival discrimination~\cite{harrell1982evaluating},
    \item calibration error (D21) based on calibration slope estimation~\cite{kuo2024ck4gen, van2019calibration},
    \item percentile-based NRI~\cite{mckearnan2018performance}.
\end{enumerate}

The evaluation pipeline first merges the generated predictions with the held-out survival outcomes, removes invalid predictions, and clips all predicted risks to the interval \([0,1]\) before downstream analysis. Calibration analysis is then performed using quantile-based risk bins and Kaplan--Meier survival estimation at the 1-year prediction horizon. Finally, percentile-based NRI is computed to quantify whether the Qwen model improves relative patient ranking compared with the Cox-derived teacher predictions.

The core components of the model evaluation framework that we focus on are summarised below:

\begin{itemize}
    \item \textbf{C-index:}\\Predicted survival risks were evaluated using pairwise survival ranking consistency between predicted risks and observed event times.
    
    \item \textbf{Calibration analysis and calibration error:}\\Predicted risks were compared against Kaplan--Meier-estimated observed risks using quantile-based calibration analysis and calibration slope estimation.
    
    \item \textbf{Percentile-based NRI:}\\Relative patient risk reordering between the Qwen and Cox models was quantified using percentile-based reclassification statistics.

\end{itemize}

%###===###%###===###%###===###%###===###%###===###
%###===###%###===###%###===###%###===###%###===###
%###===###%###===###%###===###%###===###%###===###
\newpage
\subsubsection{Details 5.1: C-index}
\label{App:CIndexEvaluation}

The concordance index evaluates whether patients with higher predicted risks tend to experience events earlier than patients with lower predicted risks. It is a rank-based discrimination metric widely used in survival analysis.

Let: $r_i$
denote the predicted risk score for patient \(i\), $T_i$ denote the observed survival duration, and $E_i \in \{0,1\}$ denote the event indicator.

The C-index is defined as:
\[
\mathrm{C\text{-}index}
=
\frac{
\sum_{i,j}
\mathbf{1}(T_i < T_j)
\mathbf{1}(r_i > r_j)
\mathbf{1}(E_i = 1)
}{
\sum_{i,j}
\mathbf{1}(T_i < T_j)
\mathbf{1}(E_i = 1)
}.
\]

Intuitively, the metric measures the proportion of comparable patient pairs for which the predicted risks are correctly ordered relative to the observed event times.

The relevant implementation is shown below:
\begin{lstlisting}[language=Python, style=mystyle, backgroundcolor=\color{backcolour4Input}]
from lifelines.utils import concordance_index

cindex = concordance_index(
    event_times=df_eval["Duration"],
    predicted_scores=-df_eval[pred_col],
    event_observed=df_eval["Event"],
)
\end{lstlisting}

The negative sign is used because larger predicted risks correspond to shorter expected survival times.

Within the experimental loop, C-index was computed separately for both the Cox teacher model and the Qwen survival model:
\begin{lstlisting}[language=Python, style=mystyle, backgroundcolor=\color{backcolour4Input}]
cox_cindex = concordance_index(
    event_times=test_df["Duration"],
    predicted_scores=-cox_partial_hazard.values.ravel(),
    event_observed=test_df["Event"],
)

qwen_cindex = concordance_index(
    event_times=test_df.loc[
        eval_df.loc[valid, "row_id"], "Duration"],
    predicted_scores=-eval_df.loc[
        valid, "qwen_pred_risk"].clip(0, 1).values,
    event_observed=test_df.loc[
        eval_df.loc[valid, "row_id"], "Event"],
)
\end{lstlisting}

This metric therefore evaluates the discriminative ability of the generated survival-risk predictions under censored time-to-event outcomes.

%###===###%###===###%###===###%###===###%###===###
%###===###%###===###%###===###%###===###%###===###
%###===###%###===###%###===###%###===###%###===###
\newpage
\subsubsection{Details 5.2: Calibration analysis and calibration error}
\label{App:CalibrationEvaluation}

Calibration analysis evaluates whether predicted survival risks numerically agree with empirically observed event probabilities.

Patients are first grouped into quantile-based risk bins. For each bin:
\begin{enumerate}
    \item the mean predicted risk is computed,
    \item the observed 1-year event probability is estimated using Kaplan--Meier survival estimation.
\end{enumerate}

Let: $\hat{r}_b$ denote the mean predicted risk in bin \(b\), and $o_b$  denote the observed Kaplan--Meier risk in the same bin. Calibration is assessed using a regression without intercept:
\[
\hat{r}_b = \beta o_b.
\]
(note that in this formulation, we are essentially putting the predicted risk on the y-axis).

The ideal calibration slope is $\beta = 1$ and thus the calibration error used throughout the framework is:
\[
D_{21} = |\beta - 1|.
\]

Observed risks are estimated using the Kaplan--Meier estimator:
\[
\hat{r}(t)
=
1 - \hat{S}(t),
\]
where
\[
\hat{S}(t)
=
\prod_{t_i \le t}
\left(
1 - \frac{d_i}{n_i}
\right).
\]

The relevant implementation is shown below:
\begin{lstlisting}[language=Python, style=mystyle, backgroundcolor=\color{backcolour4Input}]
from lifelines import KaplanMeierFitter

plot_df["risk_bin"] = pd.qcut(
    plot_df[pred_col],
    q=n_bins,
    duplicates="drop"
)

kmf = KaplanMeierFitter()

for _, g in plot_df.groupby("risk_bin", observed=False):

    kmf.fit(    g["Duration"],  event_observed=g["Event"]   )

    observed_risk = (   1.0 - float(kmf.predict(HORIZON_DAYS))  )
\end{lstlisting}

The calibration slope and calibration error are then computed using:
\begin{lstlisting}[language=Python, style=mystyle, backgroundcolor=\color{backcolour4Input}]
reg = LinearRegression(fit_intercept=False)

reg.fit(
    calib_df["observed_risk"]
        .values.reshape(-1, 1),
    calib_df["predicted_risk"].values
)

slope = float(reg.coef_[0])
calib_error = abs(slope - 1.0)
\end{lstlisting}

This calibration framework therefore measures whether the generated survival-risk predictions remain numerically consistent with empirically observed event frequencies.

%###===###%###===###%###===###%###===###%###===###
%###===###%###===###%###===###%###===###%###===###
%###===###%###===###%###===###%###===###%###===###
\newpage
\subsubsection{Details 5.3: Percentile-based NRI}
\label{App:NRIEvaluation}

The percentile-based NRI evaluates whether the Qwen survival model improves relative patient risk ranking compared with the Cox teacher model.

Let: $r_i^{\mathrm{Cox}}$
denote the Cox-derived risk for patient \(i\), and $r_i^{\mathrm{Qwen}}$
denote the corresponding Qwen-generated risk. Both risks are converted into percentile ranks:
\[
p_i^{\mathrm{Cox}}
=
\mathrm{rank}(r_i^{\mathrm{Cox}}),
\qquad
p_i^{\mathrm{Qwen}}
=
\mathrm{rank}(r_i^{\mathrm{Qwen}}).
\]

The percentile shift is:
\[
\Delta_i
=
p_i^{\mathrm{Qwen}}
-
p_i^{\mathrm{Cox}}.
\]

For event cases:
\[
\mathrm{NRI}_{\mathrm{cases}}
=
P(\Delta_i > 0 \mid \mathrm{event})
-
P(\Delta_i < 0 \mid \mathrm{event}),
\]
which rewards upward reclassification of patients who experience events.

For controls:
\[
\mathrm{NRI}_{\mathrm{controls}}
=
P(\Delta_i < 0 \mid \mathrm{no\ event})
-
P(\Delta_i > 0 \mid \mathrm{no\ event}),
\]
which rewards downward reclassification of patients without events.

The total percentile-based NRI is:
\[
\mathrm{NRI}_{\mathrm{total}}
=
\mathrm{NRI}_{\mathrm{cases}}
+
\mathrm{NRI}_{\mathrm{controls}}.
\]

The relevant implementation is shown below:
\begin{lstlisting}[language=Python, style=mystyle, backgroundcolor=\color{backcolour4Input}]
nri_df["cox_percentile"] = (
    nri_df["cox_risk"].rank(pct=True)
)

nri_df["new_percentile"] = (
    nri_df[pred_col].rank(pct=True)
)

nri_df["delta"] = (
    nri_df["new_percentile"]
    - nri_df["cox_percentile"]
)
\end{lstlisting}

The NRI statistics are then computed using:
\begin{lstlisting}[language=Python, style=mystyle, backgroundcolor=\color{backcolour4Input}]
nri_cases       = p_up_case - p_down_case
nri_controls    = p_down_control - p_up_control
nri_total       = nri_cases + nri_controls
\end{lstlisting}

The evaluation pipeline additionally removes censored patients before the prediction horizon:
\begin{lstlisting}[language=Python, style=mystyle, backgroundcolor=\color{backcolour4Input}]
nri_df = nri_df[
    ~(
        (nri_df["Event"] == 0)
        &
        (nri_df["Duration"] < horizon_days)
    )
].copy()
\end{lstlisting}

This metric therefore measures whether the Qwen survival model improves (or worsens) clinically meaningful patient risk ranking relative to the Cox-derived teacher supervision.

%###===###%###===###%###===###%###===###%###===###
%###===###%###===###%###===###%###===###%###===###
%###===###%###===###%###===###%###===###%###===###
\newpage
\section{Representation Analysis}\label{App:Rep}

\begin{algorithm}[h!]
\caption{Hidden-state extraction, t-SNE representation analysis, and SHAP explainability}
\label{alg:representation_explainability_pipeline}
\begin{algorithmic}[1]
\Require Held-out patient prompts $P$, fine-tuned Qwen model $M$, feature set $X$
\Ensure Hidden-state embeddings, low-dimensional visualisations, and SHAP feature attributions

\State Load the fine-tuned Qwen model with the trained LoRA adapter
\State Switch the model into inference mode

\For{each held-out patient prompt $p_i \in P$}
    \State Generate the autoregressive survival-risk prediction
    \State Parse the generated text into a scalar risk value
    \State Extract the final-layer hidden-state representation from the transformer
    \State Store the hidden-state embedding and associated predicted risk
\EndFor

\State Stack all hidden-state embeddings into a representation matrix
\State Apply PCA dimensionality reduction to the embedding matrix
\State Apply t-SNE to the reduced embedding representation
\State Visualise the resulting two-dimensional embedding space using risk-based colouring

\end{algorithmic}
\end{algorithm}

Following held-out inference, the framework was extended with additional representation-analysis. The first component of this extension focuses on hidden-state representation extraction. During inference, each structured patient prompt was passed through the fine-tuned Qwen model, after which the hidden-state activations from the final transformer layer were extracted. Specifically, the hidden representation corresponding to the final non-padding token was used as a compact patient-level embedding vector. These embeddings provide a learned latent representation of the patient prompt conditioned on the survival-prediction task and therefore enable downstream analysis of the model’s internal representation structure. After extracting hidden-state embeddings for all held-out patients, dimensionality-reduction analysis was performed using principal component analysis (PCA) followed by two-dimensional t-distributed stochastic neighbour embedding (t-SNE)~\cite{van2008visualizing}. 

Let $p_i$ denote the structured textual prompt corresponding to patient \(i\). After tokenisation, the autoregressive transformer produces a sequence of hidden-state activations:
\[
H_i^{(l)}
=
\left(
h_{i,1}^{(l)},
h_{i,2}^{(l)},
\dots,
h_{i,T_i}^{(l)}
\right),
\]
where: $l$ denotes the transformer layer index, $T_i$ denotes the tokenised sequence length for patient \(i\),
and $h_{i,t}^{(l)} \in \mathbb{R}^d$ represents the hidden-state vector at token position \(t\).

In this work, the representation vector for each patient was defined as the hidden-state activation corresponding to the final non-padding token from the final transformer layer:
\[
z_i
=
h_{i,T_i}^{(L)},
\]
where: $L$ denotes the final transformer layer.

The extraction of the hidden-state embedding is implemented in the following function:
\begin{lstlisting}[language=Python, style=mystyle, backgroundcolor=\color{backcolour4Input}]
def get_last_token_embedding(model, tokenizer, prompt: str) -> np.ndarray:
    
    inputs = tokenizer(
        prompt,
        return_tensors="pt",
        truncation=True,
        max_length=MAX_SEQ_LENGTH
    ).to("cuda")

    with torch.no_grad():
        outputs = model(
            **inputs,
            output_hidden_states=True,
            return_dict=True,
        )

    hidden = outputs.hidden_states[-1]

    attn = inputs["attention_mask"]
    last_idx = attn.sum(dim=1) - 1

    vec = hidden[0, last_idx.item(), :] \
        .detach() \
        .float() \
        .cpu() \
        .numpy()

    return vec
\end{lstlisting}

\underline{Lines 11--15}:\\
The key operation enabling hidden-state extraction is:
\begin{lstlisting}[language=Python, style=mystyle, backgroundcolor=\color{backcolour4Input}]
outputs = model(
    **inputs,
    output_hidden_states=True,
    return_dict=True,
)
\end{lstlisting}

which instructs the transformer to return the full sequence of hidden-state activations from all transformer layers. The final-layer activations are then obtained using:
\begin{lstlisting}[language=Python, style=mystyle, backgroundcolor=\color{backcolour4Input}]
hidden = outputs.hidden_states[-1]
\end{lstlisting}
and we identifies the final non-padding token position using the attention mask:
\begin{lstlisting}[language=Python, style=mystyle, backgroundcolor=\color{backcolour4Input}]
attn = inputs["attention_mask"]
last_idx = attn.sum(dim=1) - 1
\end{lstlisting}
to extract the corresponding hidden-state vector:
\begin{lstlisting}[language=Python, style=mystyle, backgroundcolor=\color{backcolour4Input}]
vec = hidden[0, last_idx.item(), :]
\end{lstlisting}

After fine-tuning, hidden-state embeddings were extracted for all held-out test patients using the same structured prompt format employed during training and inference:
\begin{lstlisting}[language=Python, style=mystyle, backgroundcolor=\color{backcolour4Input}]
emb_list: List[np.ndarray] = []

for _, row in test_df.iterrows():

    prompt = make_prompt(row)

    gen_text, pred = generate_risk(
        model,
        tokenizer,
        prompt
    )

    emb = get_last_token_embedding(
        model,
        tokenizer,
        prompt
    )

    emb_list.append(emb)
\end{lstlisting}

The resulting patient-level hidden-state embeddings were subsequently stacked into a representation matrix for downstream representation analysis:
\begin{lstlisting}[language=Python, style=mystyle, backgroundcolor=\color{backcolour4Input}]
embeddings = np.vstack(emb_list)
\end{lstlisting}

These hidden-state vectors therefore provide a learned latent representation of structured clinical prompts conditioned on the autoregressive survival-risk prediction objective. The extracted embeddings were subsequently used for downstream PCA, t-SNE visualisation, and linear-probe analysis of the learned representation geometry.

%###===###%###===###%###===###%###===###%###===###
%###===###%###===###%###===###%###===###%###===###
%###===###%###===###%###===###%###===###%###===###
\newpage
\section{Hyperparameters}\label{App:Hyperparameters}
\begin{table}[h!]
\small
\centering
\caption{Replication-oriented summary of experimental hyperparameters and configuration settings.}
\label{tab:experiment_hyperparameters}
\renewcommand{\arraystretch}{1.2}
\begin{tabular}{p{0.1cm} p{4cm} p{4cm} p{4cm}}
\hline
& \textbf{Hyperparameter / setting} & \textbf{Value} & \textbf{Purpose / note} \\
\hline
\hline

\multicolumn{4}{l}{\textit{Data splitting and preprocessing} (Appendix \S~\ref{App:RiskCreation})} \\
\hline
 & Train/test split & 50:50 & Half for testing. \\
 & Split stratification & Event & \\
 & Feature set & All except Duration/Event & Used for prompting and Cox\newline prediction. \\\\

\hline
\multicolumn{4}{l}{\textit{Prompting and completion formatting} (Appendix \S~\ref{App:PromptTaskInstruction})} \\
\hline
 & Prompt instruction & 1-year risk prediction & Model instructed to output\newline scalar risk in $[0,1]$. \\
 & Completion format & 4 decimal places & Fixed-length numeric target\newline string. \\\\

\hline
\multicolumn{4}{l}{\textit{Base model configuration} (Appendix \S~\ref{App:LoRALoading})} \\
\hline
 & Base model & Qwen2.5-1.5B-Instruct & \\
 & Quantisation & 4-bit &  \\
 & Maximum sequence length & 512 & \\\\

\hline
\multicolumn{4}{l}{\textit{LoRA configuration} (Appendix \S~\ref{App:LoRALoading})} \\
\hline
 & Rank ($r$) & 16 & \\
 & LoRA alpha & 16 &  \\
 & LoRA dropout & 0.05 &  \\
 & Target modules & q/k/v/o + MLP projections & \\\\

\hline
\multicolumn{4}{l}{\textit{Supervised fine-tuning configuration} (Appendix \S~\ref{App:SFTOperation})} \\
\hline
 & Batch size per device & 2 &  \\
 & Gradient accumulation & 8 & Effective batch size of 16. \\
 & Number of epochs & 10 & \\
 & Learning rate & $1\times10^{-4}$ &  \\
 & Maximum sequence length & 512 &  \\
 & Completion-only loss & True &  \\
 & FP16 training & True & \\
 & BF16 training & False & \\\\

\hline
\multicolumn{4}{l}{\textit{Inference configuration} (Appendix \S~\ref{App:HeldOutInferenceGeneration})} \\
\hline
 & Maximum generated tokens & 16 & Short deterministic completion\newline generation. \\
 & Sampling & False &  \\
 & Temperature & 0.0 & Deterministic inference. \\
 & Parsing rule & First float & \\\\
\hline
\end{tabular}
\end{table}

%%%===%%%%%%===%%%%%%===%%%%%%===%%%%%%===%%%%%%===%%%%%%===%%%%%%===%%%%%%===%%%
%%%===%%%%%%===%%%%%%===%%%%%%===%%%%%%===%%%%%%===%%%%%%===%%%%%%===%%%%%%===%%%
%%%===%%%%%%===%%%%%%===%%%%%%===%%%%%%===%%%%%%===%%%%%%===%%%%%%===%%%%%%===%%%
\newpage
\section{More Experiments}\label{App:MoreExperiments}
\subsection{GBSG2 Dataset}
\label{App:MoreDetailsGbsg2}

The German Breast Cancer Study Group 2 (GBSG2) dataset originates from the study by Schumacher \textit{et al.}~\cite{schumacher1994randomized}, which investigated prognostic factors and treatment effects among patients with node-positive breast cancer. In this work, we utilise a transformed version of the GBSG2 dataset available through the \texttt{lifelines} package. Following the methodology of Sauerbrei \textit{et al.}~\cite{sauerbrei1999modelling}, several continuous variables were categorised into clinically interpretable groups. Specifically, age was grouped into \(\leq 45\) years, 46--60 years, and \(>60\) years; tumour size into \(\leq 20\) mm, 21--30 mm, and \(>30\) mm; and the number of positive lymph nodes into \(\leq 3\), 4--9, and \(\geq 10\) nodes. Menopausal status, hormonal therapy, progesterone receptor level, and oestrogen receptor level were treated as binary variables, while tumour grade was represented using ordinal categories.

Hazard ratios obtained from the Cox proportional hazards model are presented in Table~\ref{Tab:AppGBSG2}. Several clinically meaningful associations were observed. Increasing tumour burden, reflected by larger tumour size and higher numbers of positive lymph nodes, was associated with substantially increased mortality risk. Patients with ten or more positive lymph nodes demonstrated the strongest adverse prognostic effect, with a hazard ratio of 3.50 relative to patients with three or fewer nodes. Higher tumour grade and low progesterone receptor levels were also associated with poorer survival outcomes. Conversely, hormonal therapy demonstrated a protective effect, with estimated 33\% hazard reduction.

Descriptive statistics for the cohort are summarised in Table~\ref{Tab:GBSG2}. Approximately 43.6\% of patients experienced the event of interest during follow-up, with a median follow-up duration of 1084 days. The dataset contains a balanced distribution across several clinically important characteristics, including menopausal status, tumour grade, tumour size, receptor status, and nodal involvement.

%%%===%%%%%%===%%%%%%===%%%%%%===%%%%%%===%%%%%%===%%%%%%===%%%%%%===%%%%%%===%%%
%%%===%%%%%%===%%%%%%===%%%%%%===%%%%%%===%%%%%%===%%%%%%===%%%%%%===%%%%%%===%%%
%%%===%%%%%%===%%%%%%===%%%%%%===%%%%%%===%%%%%%===%%%%%%===%%%%%%===%%%%%%===%%%
%\newpage
\subsection{ACTG320 Dataset}
\label{App:MoreDetailsACTG320}

The ACTG320 dataset originates from the study by Hammer \textit{et al.}~\cite{hammer1997controlled}, which investigated the efficacy of adding the protease inhibitor indinavir to a two-drug antiretroviral regimen in HIV-1 infected patients with prior zidovudine therapy and CD4 counts below 200 cells/mm\(^3\). In this work, we utilise a variation of the ACTG320 dataset available through the \texttt{sksurv} package. The dataset includes demographic information, treatment allocation, immunological status, functional impairment, prior treatment exposure, and time-to-event outcomes, where the event corresponds to the development of an AIDS-defining condition or death.

To increase modelling flexibility while preserving clinical interpretability, we included six variables in the Cox proportional hazards model: age, sex, CD4 cell count stratum, treatment indicator, functional impairment, and months of prior zidovudine (ZDV) use. CD4 cell count was categorised into \(\leq 50\) cells/mm\(^3\) and 51--200 cells/mm\(^3\). Functional impairment was derived from the Karnofsky Performance Scale (KPS)~\cite{karnofsky1948use}, where increasing values correspond to greater impairment severity. Specifically, KPS scores of 100, 90, 80, and 70 were mapped to impairment levels of 0, 1, 2, and 3 respectively.

Hazard ratios obtained from the Cox proportional hazards model are presented in Table~\ref{Tab:AppACTG320}. Several clinically meaningful associations were observed. Patients receiving the treatment regimen demonstrated substantially lower mortality risk compared with the control group, with a hazard ratio of 0.51. Higher CD4 cell counts were also associated with improved survival outcomes. In contrast, increasing functional impairment was strongly associated with elevated hazard, nearly doubling the estimated risk for each increase in impairment level. Age showed a modest increase in hazard, while sex and prior ZDV exposure demonstrated limited independent prognostic contribution.

Descriptive statistics for the cohort are summarised in Table~\ref{Tab:ACTG320}. Approximately 8.3\% of patients experienced the event of interest during follow-up, with a median follow-up duration of 257 days. The cohort was predominantly male and contained a balanced distribution between treatment groups and CD4 count strata. 

%%%===%%%%%%===%%%%%%===%%%
%%%===%%%%%%===%%%%%%===%%%
%%%===%%%%%%===%%%%%%===%%%
\newpage
\begin{table}[h!]
\footnotesize
\centering
\begin{tabular}{|p{6cm}|p{5.5cm}|}
\hline
\textbf{Variable} & \textbf{HR (95\% CI)\newline (Ours)} \\
\hline
\hline
\textbf{Age} & \\
\(\leq 45\) years & Baseline \\
46-60 years & 0.64 (0.45, 0.91) \\
\(>60\) years & 0.65 (0.41, 1.03) \\
\hline
\hline
\textbf{Menopausal State} & \\
Pre-menopausal & Baseline \\
Post-menopausal & 1.31 (0.94, 1.83) \\
\hline
\hline
\textbf{Tumour Size} & \\
\(\leq 20\) mm & Baseline \\
21-30 mm & 1.22 (0.90, 1.65) \\
\(>30\) mm & 1.31 (0.95, 1.79) \\
\hline
\hline
\textbf{Tumour Grade} & \\
I & Baseline \\
II & 1.74 (1.06, 2.85) \\
III & 1.75 (1.02, 3.02) \\
\hline
\hline
\textbf{Number of Positive Nodes} & \\
\(\leq 3\) & Baseline \\
4-9 & 1.97 (1.51, 2.58) \\
\(\geq 10\) & 3.50 (2.57, 4.74) \\
\hline
\hline
\textbf{Progesterone Receptor Level} & \\
\(\geq 20\) fmol/mg & Baseline \\
\(<20\) fmol/mg & 1.85 (1.39, 2.45) \\
\hline
\hline
\textbf{Oestrogen Receptor Level} & \\
\(\geq 20\) fmol/mg & Baseline \\
\(<20\) fmol/mg & 1.00 (0.77, 1.33) \\
\hline
\hline
\textbf{Hormonal Therapy} & \\
False & Baseline \\
True & 0.67 (0.52, 0.86) \\
\hline
\end{tabular}
\caption{\label{Tab:AppGBSG2}Hazard ratios obtained from the Cox model fitted to the GBSG2 dataset.}
\end{table}

%###===>>>%###===>>>%###===>>>%###===>>>%###===>>>%###===>>>
%###===>>>%###===>>>%###===>>>%###===>>>%###===>>>%###===>>>
\begin{table}[h!]
\footnotesize
\centering
\begin{tabular}{|p{3cm}|p{2cm}|p{3cm}|p{4cm}|}
\hline
\textbf{Variable} & \textbf{Type} & \textbf{Category/Statistic} & \textbf{Value/Proportion} \\
\hline
\hline
\textbf{Age} & Categorical & \(\leq 45\) years \newline 46-60 years \newline \(> 60\) years & Baseline \newline 53.06\% \newline 28.86\% \\
\hline
\hline
\textbf{Menopausal\newline State} & Binary & Pre- \newline Post- & Baseline \newline 53.79\% \\
\hline
\hline
\textbf{Tumour Size} & Categorical & \(\leq 20\) mm \newline 21-30 mm \newline \(> 30\) mm & Baseline \newline 43.00\% \newline 31.20\% \\
\hline
\textbf{Tumour Grade} & Categorical & I \newline II \newline III & Baseline \newline 67.2\% \newline 21.28\% \\
\hline
\hline
\textbf{Number of\newline Positive Nodes} & Categorical & \(\leq 3\) \newline 4-9 \newline \(\geq 10\) & Baseline \newline 30.17\% \newline 14.72\% \\
\hline
\hline
\textbf{Progesterone\newline Receptor Level} & Binary & \(\geq 20\) fmol/mg \newline \(< 20\) fmol/mg & Baseline \newline 37.46\% \\
\hline
\textbf{Oestrogen\newline Receptor Level} & Binary & \(\geq 20\) fmol/mg \newline \(< 20\) fmol/mg & Baseline \newline 37.76\% \\
\hline
\hline
\textbf{Hormonal Therapy} & Binary & False \newline True & Baseline \newline 33.97\% \\
\hline
\hline
\textbf{Event} & Binary & Occurred & 43.59\% \\
\hline
\textbf{Duration} & Numeric & Mean ± Std Dev & 1124.49 ± 642.79 \\
\hline
\end{tabular}
\caption{\label{Tab:GBSG2}An overview of the structure and descriptive statistics for the GBSG2 synthetic dataset.}
\end{table}

%%%===%%%%%%===%%%%%%===%%%
%%%===%%%%%%===%%%%%%===%%%
%%%===%%%%%%===%%%%%%===%%%
\newpage
\begin{table}[h!]
\footnotesize
\centering
\begin{tabular}{|p{6cm}|p{5.5cm}|}
\hline
\textbf{Variable} & \textbf{HR (95\% CI)\newline (Ours)} \\
\hline
\hline
\textbf{Age} & 1.02 (1.00, 1.04) \\
\hline
\hline
\textbf{Sex} & \\
Female & Baseline \\
Male & 0.93 (0.53, 1.62) \\
\hline
\hline
\textbf{CD4 Cell Count} & \\
\(\leq 50\) cells/mm\(^3\) & Baseline \\
51--200 cells/mm\(^3\) & 0.41 (0.26, 0.64) \\
\hline
\hline
\textbf{Treatment Indicator} & \\
Control group & Baseline \\
Treatment group & 0.51 (0.33, 0.77) \\
\hline
\hline
\textbf{Functional Impairment} & 1.93 (1.53, 2.44) \\
\hline
\hline
\textbf{Months of Prior ZDV Use} & 1.00 (0.99, 1.01) \\
\hline
\end{tabular}
\caption{\label{Tab:AppACTG320}Hazard ratios obtained from the Cox model fitted to the ACTG320 dataset.}
\end{table}

%###===>>>%###===>>>%###===>>>%###===>>>%###===>>>%###===>>>
%###===>>>%###===>>>%###===>>>%###===>>>%###===>>>%###===>>>
\begin{table}[h!]
\footnotesize
\centering
\begin{tabular}{|p{3cm}|p{2cm}|p{3cm}|p{4cm}|}
\hline
\textbf{Variable} & \textbf{Type} & \textbf{Category/Statistic} & \textbf{Value/Proportion} \\
\hline
\hline
\textbf{Age} & Numeric & Mean ± Std Dev & 38.75 ± 9.16 \\
\hline
\hline
\textbf{Sex} & Binary & Female \newline Male & Baseline \newline 85.66\% \\
\hline
\hline
\textbf{CD4 Cell Count} & Binary & \(\leq 50\) cells/mm\(^3\) \newline 51--200 cells/mm\(^3\) & Baseline \newline 52.56\% \\
\hline
\hline
\textbf{Treatment Indicator} & Binary & Control group \newline Treatment group & Baseline \newline 49.70\% \\
\hline
\hline
\textbf{Functional\newline Impairment} & Numeric & Mean ± Std Dev & 0.86 ± 0.75 \\
\hline
\hline
\textbf{Months of\newline Prior ZDV Use} & Numeric & Mean ± Std Dev & 29.96 ± 28.49 \\
\hline
\hline
\textbf{Event} & Binary & Occurred & 8.34\% \\
\hline
\textbf{Duration} & Numeric & Mean ± Std Dev & 230.18 ± 89.88 \\
\hline
\end{tabular}
\caption{\label{Tab:ACTG320}An overview of the structure and descriptive statistics for the ACTG320 synthetic dataset.}
\end{table}

%###===###%###===###%###===###%###===###%###===###
%###===###%###===###%###===###%###===###%###===###
%###===###%###===###%###===###%###===###%###===###
\newpage
\subsection{Performance Evaluation on GBSG2 \& ACTG320}\label{App:MoreResults}
\begin{table}[h!]
\centering
\footnotesize
\caption{
Comparison of the Cox teacher and Qwen-based LLM student on GBSG2.}
\label{tab:gbsg2_combined}
\begin{tabular}{lcc}
\toprule
\multicolumn{3}{c}{\textbf{Cohort-level performance}} \\
\midrule
Metric & Cox & Qwen \\
\midrule

C-index $\uparrow$ &
$0.6780 \pm 0.0210$ &
$0.6801 \pm 0.0245$ \\

Calibration error (D21) $\downarrow$ &
$0.0996 \pm 0.0592$ &
$0.0924 \pm 0.0974$ \\

\midrule
\multicolumn{3}{c}{\textbf{Subgroup calibration error (D21) $\downarrow$}} \\
\midrule

Age 46--60y &
$0.134 \pm 0.095$ &
$0.130 \pm 0.118$ \\

Age $>60$y &
$0.094 \pm 0.038$ &
$0.081 \pm 0.059$ \\

Grade II &
$0.096 \pm 0.058$ &
$0.091 \pm 0.057$ \\

Grade III &
$0.166 \pm 0.144$ &
$0.208 \pm 0.136$ \\

Menopausal &
$0.082 \pm 0.116$ &
$0.079 \pm 0.132$ \\

Nodes 4--9 &
$0.144 \pm 0.077$ &
$0.145 \pm 0.042$ \\

Nodes $\geq 10$ &
$0.068 \pm 0.071$ &
$0.063 \pm 0.076$ \\

Oest $<20$ fmol &
$0.127 \pm 0.097$ &
$0.154 \pm 0.088$ \\

Prog $<20$ fmol &
$0.164 \pm 0.074$ &
$0.136 \pm 0.097$ \\

Therapy &
$0.079 \pm 0.077$ &
$0.089 \pm 0.088$ \\

Tumor 21--30mm &
$0.085 \pm 0.055$ &
$0.091 \pm 0.100$ \\

Tumor $>30$mm &
$0.171 \pm 0.116$ &
$0.194 \pm 0.146$ \\

\bottomrule
\end{tabular}
\end{table}

\begin{table}[h!]
\centering
\footnotesize
\caption{
Comparison of the Cox teacher and Qwen-based LLM student on ACTG320.}
\label{tab:actg320_combined}
\begin{tabular}{lcc}
\toprule
\multicolumn{3}{c}{\textbf{Cohort-level performance}} \\
\midrule
Metric & Cox & Qwen \\
\midrule

C-index $\uparrow$ &
$0.7038 \pm 0.0239$ &
$0.7066 \pm 0.0247$ \\

Calibration error (D21) $\downarrow$ &
$0.1532 \pm 0.1154$ &
$0.1809 \pm 0.1973$ \\

\midrule
\multicolumn{3}{c}{\textbf{Subgroup calibration error (D21) $\downarrow$}} \\
\midrule

CD4 51--200 &
$0.255 \pm 0.044$ &
$0.265 \pm 0.135$ \\

Sex (male) &
$0.173 \pm 0.128$ &
$0.187 \pm 0.156$ \\

Treatment &
$0.347 \pm 0.230$ &
$0.416 \pm 0.261$ \\

\bottomrule
\end{tabular}
\end{table}

Table~\ref{tab:gbsg2_combined} summarises the held-out performance on GBSG2 (reported in mean $\pm$ std). The Qwen-based LLM student achieves performance highly comparable to the Cox teacher, slightly exceeding the Cox model in mean C-index ($0.6801$ versus $0.6780$) while also obtaining marginally lower overall calibration error ($0.0924$ versus $0.0996$). Subgroup calibration analysis demonstrates broadly consistent behaviour between the two models across most clinically relevant categories. In several subgroups, including Age $>60$y, Grade II, Menopausal status, and Nodes $\geq 10$, the LLM student achieves slightly lower D21 calibration error than the Cox model. However, calibration degradation is observed in some higher-risk subgroups, particularly Grade III disease and Tumor $>30$mm, where the Qwen model exhibits moderately higher calibration error.

Table~\ref{tab:actg320_combined} summarises the held-out performance on ACTG320. Similar to the GBSG2 results, the Qwen-based LLM student achieves discrimination performance close to the Cox teacher, with a slightly higher mean C-index ($0.7066$ versus $0.7038$). However, calibration performance is less stable, with the Qwen model showing higher overall D21 calibration error ($0.1809$ versus $0.1532$) and substantially larger variability across runs. Subgroup calibration analysis reveals broadly similar behaviour between the models for CD4 and sex-based subgroups, although the Qwen student consistently exhibits slightly higher calibration error. The largest discrepancy is observed in the treatment subgroup, where the LLM student demonstrates noticeably poorer calibration than the Cox teacher ($0.416$ versus $0.347$), suggesting that treatment-related risk stratification may be more difficult for the distilled model to preserve reliably.

%###===###%###===###%###===###%###===###%###===###
%###===###%###===###%###===###%###===###%###===###
%###===###%###===###%###===###%###===###%###===###
%\newpage
%\input{checklist_My.tex}

\end{document}